\documentclass[10pt,runningheads,a4paper]{llncs}
\usepackage{amssymb}
\usepackage{graphicx}
\usepackage{hyperref}
\usepackage{algorithm,float}
\usepackage[noend]{algpseudocode}
\usepackage{cite}
\usepackage[numbers]{natbib}
\usepackage{longtable}
\usepackage{booktabs, makecell, longtable}
\makeatletter
\def\BState{\State\hskip-\ALG@thistlm}
\usepackage[margin=1 in]{geometry}
\usepackage{booktabs}
\usepackage{multirow}
\usepackage[table,xcdraw]{xcolor}
\usepackage{amssymb}% http://ctan.org/pkg/amssymb
\usepackage{pifont}% http://ctan.org/pkg/pifont
\usepackage{amsmath}
\usepackage{pdfpages}
\usepackage{url}

\urldef{\mailsa}\path|{alfred.hofmann, ursula.barth, ingrid.haas, frank.holzwarth,|
	\urldef{\mailsb}\path|anna.kramer, leonie.kunz, christine.reiss, nicole.sator,|
	\urldef{\mailsc}\path|erika.siebert-cole, peter.strasser, lncs}@springer.com|    
\newcommand{\keywords}[1]{\par\addvspace\baselineskip	\noindent\keywordname\enspace\ignorespaces#1}
\newcommand{\xmark}{\ding{55}}%
\begin{document}
	
\raggedbottom	
	\mainmatter 
	
\title{Deep Learning based Computer-Aided Diagnosis Systems for Diabetic Retinopathy: A Survey}
\titlerunning{Deep Learning based Computer-Aided Diagnosis Systems for Diabetic Retinopathy: A Survey}	
\author{Norah Asiri\inst{1,}\inst{2}, Muhammad Hussain\inst{2}, Fadwa Al Adel\inst{3},Nazih Alzaidi\inst{4}}
\authorrunning{Norah Asiri et al.}
\institute{Community College, Imam Abdulrahman Bin Faisal University, Dammam, Saudi Arabia\inst{1}\\Computer and Information Science College, King Saud University, Riyadh, Saudi Arabia\inst{2}\\Department of Ophthalmology, College of Medicine, Princess Nourah bint Abdulrahman University, Riyadh, Saudi Arabia\inst{3}\\Vitreoretinal Diseases and Surgery, King Khalid Eye Specialist Hospital, Riyadh, Saudi Arabia\inst{4}\\ \email{norah.m.asiri@outlook.com} \\ \email{mhussian@ksu.edu.sa}\\\email{ffaladel@pnu.edu.sa}\\\email{nzaidi@kkesh.med.sa}}
	\maketitle

\begin{abstract}
		
Diabetic retinopathy (DR) results in vision loss if not treated early. A computer-aided diagnosis (CAD) system based on retinal fundus images is an efficient and effective method for early DR diagnosis and assisting experts. A computer-aided diagnosis (CAD) system involves various stages like detection, segmentation and classification of lesions in fundus images. Many traditional machine-learning (ML) techniques based on hand-engineered features have been introduced. The recent emergence of deep learning (DL) and its decisive victory over traditional ML methods for various applications motivated the researchers to employ it for DR diagnosis, and many deep-learning-based methods have been introduced. In this paper, we review these methods, highlighting their pros and cons. In addition, we point out the challenges to be addressed in designing and learning about efficient, effective and robust deep-learning algorithms for various problems in DR diagnosis and draw attention to directions for future research.
				
\end{abstract}

\section{Introduction}
Diabetic retinopathy (DR) is one of the main causes of blindness among the working-age population. It is one of the most feared complications of diabetes. The fundamental problem of DR is that it becomes incurable at advanced stages, so early diagnosis is important. However, this involves remarkable difficulty in the health care system due to a large number of potential patients and the small number of experienced technicians. This has motivated the need to develop automated diagnosis systems to assist in early diagnosis of DR. Several attempts have been made in this direction, and several approaches based on hand-engineered features have been proposed, which have shown promising efficiency in recognizing DR regions in retinal fundus images. 

Hand-engineered features are commonly used with traditional machine-learning (ML) methods for DR diagnosis. Different surveys have reviewed these traditional methods   \cite{mookiah2013computer,faust2012algorithms,joshi2018review,mansour2017evolutionary,almotiri2018retinal,almazroa2015optic,thakur2018survey}. For example, \citet{mookiah2013computer,mansour2017evolutionary} categorized DR diagnosis according to the adopted methodologies, such as mathematical morphology, retinal lesion tracking, thresholding and deformable models, clustering-based models, matched filtering models and hybrid approaches. \citet{faust2012algorithms} reviewed algorithms that extract lesion features from fundus images, such as the blood vessel area, exudes, hemorrhages, microaneurysms and texture. \citet{joshi2018review} reviewed the early research on exudate detection. \citet{almotiri2018retinal} provided an overview of algorithms to segment retinal vessels. \citet{almazroa2015optic} and \citet{thakur2018survey} reviewed several methods for optic disc segmentation and diagnosis of glaucoma. However, expert knowledge is a prerequisite for hand-engineered features, and choosing the appropriate features requires intensive investigation of various options and tedious parameter settings. Moreover, techniques based on hand-engineered features do not generalize well.	

In recent years, the availability of huge datasets and the tremendous computing power offered by graphics processing units (GPUs) have motivated research on deep-learning algorithms, which have shown outstanding performance in various computer vision tasks and have gained a decisive victory over traditional hand-engineered-based methods. Many deep-learning (DL)-based algorithms have also been developed for various tasks to analyze retinal fundus images to develop automatic computer-aided diagnosis systems for DR. This paper reviews the latest DL algorithms used in DR detection, highlighting the contributions and challenges of recent research papers. First, we provide an overview of various DL approaches and then review the DL-based techniques for DR diagnosis. Finally, we summarize future directions, gaps and challenges in designing and training deep neural networks for DR diagnosis. The remainder of the paper is organized as follows: automatic detection of DR, types of lesions, DR stages, grading of DR, detection tasks and the detection framework are presented in Section \ref{AD}. After that, public-domain DR datasets and common performance metrics are briefly described in Section \ref{DS}. An overview of DL techniques used in DR diagnosis is given in Section \ref{Deep learning}. The most recent research based on DL for DR diagnosis are reviewed in Section \ref{LS}. This research is discussed in Section \ref{DIS}. Finally, research gaps and future directions with conclusion are presented in Sections \ref{GAP} and \ref{CON}.

\section{Automatic Diabetic Retinopathy Detection}\label{AD}

In this section, for the sake of clarity, we give an overview of DR detection, types of DR lesions, stages of DR, grading of DR, DR-detection tasks and the general framework for detection. Automatic computer-aided solutions for DR characterization are still an open field of research \cite{mansour2017evolutionary}. Automatic image-based DR detection systems are intended to perform rapid retinal evaluations and early detection of DR to indicate whether DR complications are present.

\subsection{Types of Lesions} \label{TL} The earliest clinical signs of DR and retinal damage are $\textbf{microaneurysms (MAs)}$, which are a dilation of  microvasculature formed due to disruption of the internal elastic lamina.  Retinal microaneurysms reduce vision due to local loss of endothelial barrier function, causing leakage and retinal edema. MAs are small (usually less than  125 microns in diameter) and appear as red spots with sharp margins. When walls of weak capillaries are broken, bleeding causes $\textbf{hemorrhages (HMs)}$, which are similar to MAs but larger \cite{early1991grading} and have an irregular margin, they have different appearances according to which retinal layer they leak in. Splinter hemorrhages occur in the superficial surface layers of the retina and cause a superficial flame-shaped bleeding. Whereas dot and blot hemorrhages occur in the deeper layers of the retina. More leakage of damaged capillaries can cause $\textbf{exudates (EXs)}$, which usually appear  yellow and irregularly shaped in the retina. There are two types of EXs: hard and soft. $\textbf{Hard exudates (HEs)}$ are lipoproteins and other proteins escaping from abnormal retinal vessels. They are white or white-yellow with sharp margins. They are often organized in blocks or circular rings \cite{harney2006diabetic} and are located in the outer layer of the retina. On the other hand, $\textbf{soft exudates (SEs)}$ or cotton wool spots (CWS) are small, whitish-grey cloud-like shapes that occur when an arteriole is occluded \cite{mcleod2005cotton}. EXs are different from MAs and HMs in terms of brightness. MAs and HMs are dark lesions, while EXs are bright \cite{akram2014detection}. Variations in the diameter of the retinal veins is called $\textbf{Venous beading (VB)}$ \cite{lee1994parallel} this usually happens in advanced stages of non-proliferative diabetic retinopathy. Due to the inability to use glucose by normal routes, alternate blood pathways are activated, which causes the synthesis of elements such as sorbitols and favors the development of alterations in the microvasculature. $\textbf{Intraretinal microvascular abnormalities (IRMA)}$ is an example, it represents either a dilation of pre-existing capillaries or an actual growth of new blood vessels within the retina. When the retinal vessels stand out and grow towards the vitreous they are called $\textbf{neovascularization (NV)}$ \cite{patz1980studies}. $\textbf{Macular edema (ME)}$ occurs when the retinal capillaries become permeable and leakage occurs around macula \cite{dr}. This can lead to retinal thickening or hard exudates developing either within one disk diameter of the center of the macula (the fovea) \cite{early1987treatment} or involving the fovea, which is responsible for the central vision.

\begin{figure}[H]
		\centering
		\includegraphics[width=0.7\linewidth]{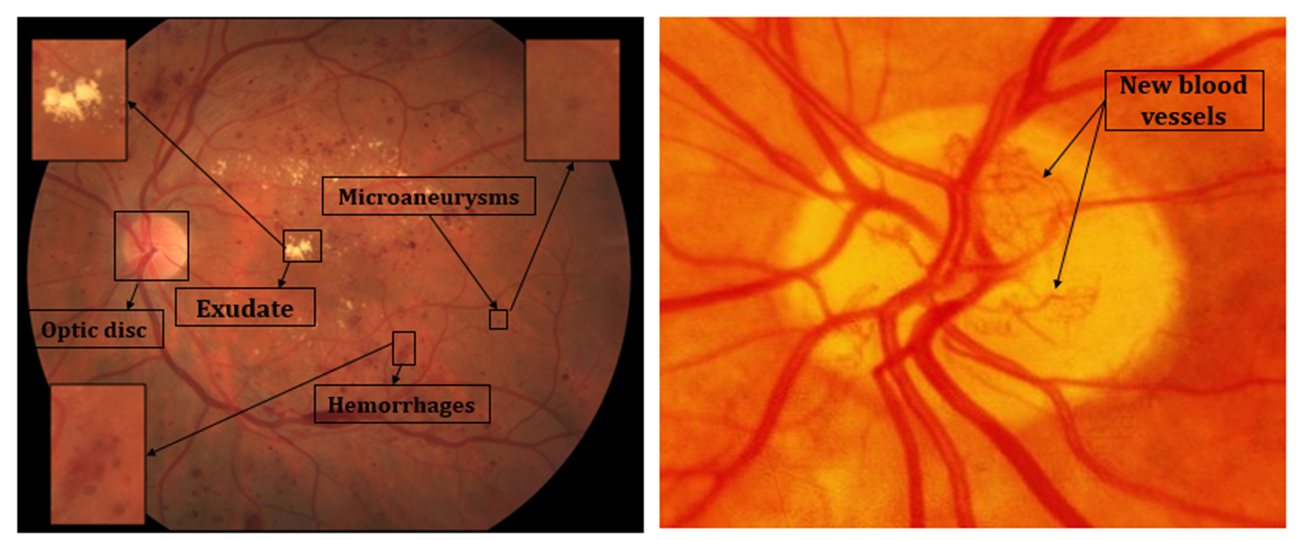}
		\caption{Optic disc and abnormal findings in the eye fundus caused by the diabetic retinopathy.$\textit{Left:}$ MA, EX and HM. $\textit{Right:}$ new blood vessel routes(PDR). }
		\label{fig:screenshot019}
	\end{figure}

An important object that plays an essential role in detecting DR is the $\textbf{optic disc (OD)}$, which characterized by the highest contrast between the circular-shaped regions \cite{sopharak2008automatic}. The optic disc is used as a landmark and frame of reference to diagnose serious eye pathologies such as glaucoma, optic disc pit, optic disc drusen and to check for any neovascularization at the disc \cite{jonas1988optic,joshi2011optic}. The OD is also used to pinpoint other structures such as the fovea. In normal retina, the edges of the OD are clear and well-defined, as shown in Figure \ref{fig:screenshot019}.

\subsection{Stages of DR}
DR can be classified into two main classes based on its severity: non-proliferative $\textbf{(NPDR)}$ and proliferative $\textbf{(PDR)}$ \cite{early1991grading,acharya2009computer}. NPDR is an early stage, during which diabetes starts to damage small blood vessels within the retina; it is very common in people with diabetes  \cite{dr}. These vessels start to discharge fluid and blood, causing the retina to swell. As time passes, the swelling or edema thickens the retina, causing blurry vision. The clinical feature of this stage is at least one microaneurysm or hemorrhage with or without hard exudates \cite{fleming2010role}. Proliferative DR is an advanced stage that leads to the growth of new blood vessels; as such, it is characterized by by abnormal vascular proliferation within the retina towards the vitreous cavity. These fragile new blood vessels can bleed into the vitreous cavity and cause severe visual loss due to vitreous hemorrhage. They can also further cause traction on the retina as they usually grow with a fibro vascular network around them that may lead to tractional retinal detachment.

\subsection{Grading of DR}

\begin{figure}
	\centering
	\includegraphics[width=0.4\linewidth]{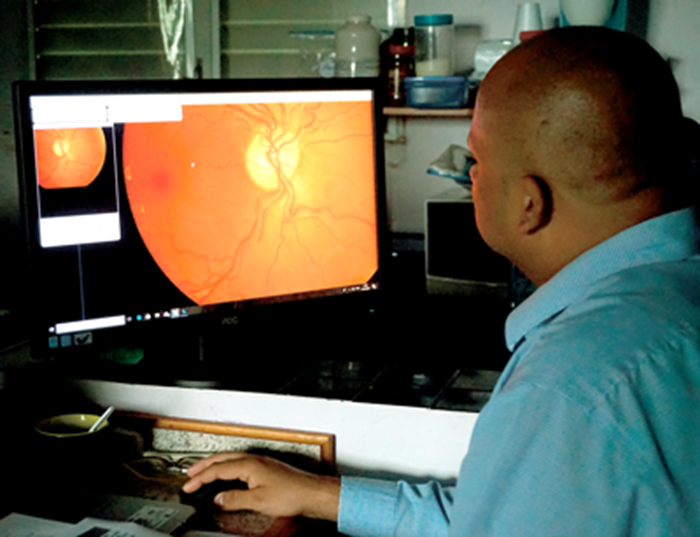}
	\caption{Graders need appropriate environment to maintain high-quality performance.}
	\label{fig:grade}
\end{figure}

Examination and screening of the retina by ophthalmoscopy usually requires dilated pupils, a skilled examiner and a visit to an eye care provider such as  optometrist to grade and classify pathology  \cite{dr_grading}.
Grading is a vital activity in DR screening programme to diagnose retinal diseases. It is an intensive procedure that needs a trained workforce and an adequately sized computer screens\ref{fig:grade}.

Graders such as optometrists or well-trained technicians perform an essential task to treat and recover potentially blinding eye conditions to treat and recover potentially blinding eye conditions such as age-related macular degeneration (AMD) and diabetic eye diseases \cite{dr_grading}.
Non-mydriatic fundus images are usually acquired but if the image is unclear due to any media opacity then mydriatic drops are used to dilate the pupil in an attempt to improve the quality of the image. All graders must receive a special training based on screening protocol to ensure the fundus images are graded in standardized manner. They should spend time on training to identify and confirm cases as having pathology abnormality or not and differentiate the levels of pathology seen and make referral decision or return for recall based on agreed interval. There are various systems to grade DR vascular changes such as American academy of ophthalmology (AAO), the classification which was introduced by the early treatment of diabetic retinopathy study (ETDRS) \cite{kanski2009clinical} and Scottish DR grading protocol where only one field is taken per eye, which is centered on the fovea \cite{zachariah2015grading}. Scottish protocol is represented in Table \ref{dr-grades}.

\begin{table}[H]
 	\centering
 			\caption{DR Scottish grading protocol\cite{dr-grades}}
 					\label{dr-grades}
 	\begin{tabular}{|l|l|l|}
 		\hline
 		Grade & Features                                                                                                                                                                & Decision               \\ \hline
 	$\textbf{R0:}$ No DR    & No abnormalities                                                                                                                                                            & Rescreen in 12 months \\ \hline
 	$\textbf{R1:}$ Mild NPDR &Only MAs & Rescreen in 12 months \\ \hline
			
 	$\textbf{R2:}$	Moderate NPDR& \begin{tabular}[c]{@{}l@{}}More than just MAs but less than severe NPDR\end{tabular}                                                            & Rescreen in 6 months  \\ \hline
 		$\textbf{R3:}$ Severe NPDR&\begin{tabular}[c]{@{}l@{}}-More than 20 HMs in each quadrant\\-Venous beading in two quadrants\\-Intraretinal microvascular abnormalities \end{tabular}                                                              & Refer                 \\ \hline
 	$\textbf{R4:}$	PDR    &\begin{tabular}[c]{@{}l@{}} -Any new vessels at OD or elsewhere\\ -Vitreous/ pre-retinal HM \end{tabular}                                                                                                                                              & Refer                 \\ \hline
 	$\textbf{M0:}$	No ME    & No EX or retinal thickening in posterior  pole                                                                                                                         & 12 month rescreening  \\ \hline
 		$\textbf{M1:}$ Mild ME    & EXs or retinal thickening at posterior pole, $>$1 disc diameters from fovea 
                                                                                                                    & 6 month rescreening   \\ \hline
 	$\textbf{M2:}$ Moderate ME & \begin{tabular}[c]{@{}l@{}}Same signs of mild ME but with 1 disc diameters  or less from fovea,\\but not affecting
 	fovea\end{tabular}                                                              & Refer for laser treatment                 \\ \hline
		
	 		$\textbf{M3:}$	Severe ME & EXs or retinal thickening affecting center of fovea                                                              & Refer for laser treatment                 \\ \hline

	 \end{tabular}
 \end{table}

	\subsection{Detection Tasks and General Framework}
At a high level, DR detection is categorized into two tasks: lesion-level-based detection and image-level-based detection. In lesion-level-based detection, every lesion is detected and their locations are determined because the number of lesions and their locations are crucial to assessing the severity level of DR \cite{seoud2016red}. On the other hand, image--based detection focuses on assessment based on image levels and is more interesting from the screening point of view because it evaluates only whether there are signs of DR \cite{seoud2016red}. Lesion-based detection usually involves two phases: (i) lesion detection and/or segmentation and (ii) lesion classification. First, lesions such as microaneurysms, hemorrhages, hard exudates and soft exudates are detected from fundus images, and the exact area of the lesion is localized. This is a challenging task because retinal fundus images contain other objects with similar appearances, such as red dots and blood vessels. For this task, the global and local context are usually needed to perform accurate localization and segmentation. The detection phase yields potential regions of interest, but they include false positives as well. The lesion-classification phase is used to remove false positives. Image-based detection is an image-screening task that classifies a given fundus image as being normal or having DR signs. This is one of the first areas of medical diagnosis to which DL has made a significant contributions \cite{litjens2017survey}.

The general framework for detection, segmentation and classification involves the specific steps of preprocessing, feature extraction/selection, choice of a suitable classification method and finally assessment of the results. DR classification systems can be divided into two types according to learning procedure: supervised and unsupervised learning. In supervised learning, the system is taught using labeled data to infer functional mapping \cite{Gondaletal17,li2016cross}. On the other hand, unsupervised learning methods tend to discover hidden patterns on their own from the properties of the unlabeled examples according to their similarity \cite{arunkumar2017multi}. Unlike hand-engineered feature-based approaches, DL approaches integrate all of the steps into a unified framework and automatically learns the features and trains the system in an end-to-end manner.

\section{Datasets and Performance Metrics}\label{DS}

In this section, we give an overview of the benchmark datasets and performance metrics that are commonly used for DR research.
\subsection{Retinal Fundus Image Datasets}

Several datasets consisting of retinal fundus images have been produced to teach and test the algorithms for different DR detection tasks. In the following paragraphs, we give an overview of the following public domain benchmark datasets: MESSIDOR \cite{messidor}, e-ophtha \cite{optha}, Kaggle \cite{kgl}, DRIVE \cite{drive}, STARE \cite{hoover2000locating}, DIARETDB1  \cite{kalviainen2007diaretdb1,dia}, CHASE \cite{chase} , DRiDB \cite{prentasic2013diabetic}, ORIGA \cite{zhang2010origa}, SCES \cite{sng2012determinants} , AREDS \cite{nih}, REVIEW \cite{al2008reference}, EyePACS-1 \cite{eyepacs}, RIM-ONE \cite{fumero2011rim}, DRISHTI-GS \cite{sivaswamy2014drishti}, ARIA \cite{aria}, DRIONS-DB \cite{drion} and SEED-DB \cite{seed}. Table \ref{dstbl} summarizes these datasets.

\subsubsection{MESSIDOR}
 It was developed under MESSIDOR research program funded by the French ministry of research and defense  \cite{messidor}. It was acquired by three ophthalmology departments using colored video 3CCD camera mounted on a Topcon TRC NW6 non-mydriatic retinograph with a 45$^{\circ}$ field of view (FOV). Two types of image level annotation were provided by expert ophthalmologists: DR grades and risk levels of macular edema. The DR grades are as follows: 

 \begin{itemize}
 
\item 0: No risk: (\#MA = 0) AND (\#HM = 0)
\item 1: (0 $<$ \#MA $\leq$ 5) AND (\#HM = 0)
\item 2: ((5 $<$ \#MA $<$ 15) OR (0 $<$ \#HM $<$ 5)) AND (NV = 0)
\item 3: (\#MA $\leq$ 15) OR (\#HM $\leq$ 5) OR (NV = 1) 	 
 \end{itemize}

The risk levels of macular edema are as follows :
 \begin{itemize}
\item 0: No risk
\item 1: Shortest distance between macula and hard EX $>$ one papilla diameter
\item 2: Shortest distance between macula and hard EX $\leq$ one papilla diameter
 \end{itemize}
 \subsubsection{e-ophtha}
 
It was introduced by e-ophtha project funded by the French research agency \cite{optha}. It provides the locations of MAs and EXs, which were identified by two ophthalmologists. The first ophthalmologist outlined the locations, which were checked and examined by the second ophthalmologist. The database consists of two datasets: e-ophtha EX and e-ophtha MA. The e-ophtha EX set contains 47 images with 12,278 EXs and 35 healthy images. Several images of healthy controls contain structures which can easily mislead EX detection methods, such as reflections and optical artifacts. On the other hand, e-ophtha MA contains 148 images with 1306 MAs and 233 healthy images.
 
\subsubsection{Kaggle} 
It consists of a large set of high-resolution retinal images taken under different conditions and was  provided by EyePACS clinics \cite{kgl}. The image level annotation was provided by expert ophthalmologists, and each image has been assigned a DR grade on the scale of 0 to 4 as follows:
\begin{itemize}

\item 0: No risk
\item1: Mild
\item 2: Moderate
\item 3: Severe
\item 4: PDR	 
\end{itemize}
 
 \subsubsection{DRIVE}
Digital retinal images for vessel extraction (DRIVE) \cite{drive} was collected under DR screening program in Netherlands for comparative studies of vascular segmentation in retinal images using Canon CR5 non-mydriatic 3CCD camera . It consists of 40 fundus images, which were randomly selected; among them 33 do not show any sign of DR whereas 7 show signs of mild early DR; it is divided into a test and training sets, each containing 20 images. It provides pixel level annotation; a pixel is annotated as a vessel pixel with 70\% confidence. %It does not contain pathology grading.

\subsubsection{STARE}

Structured analysis of the retina (STARE) \cite{hoover2000locating} program was funded by the U.S. national institutes of health (NIH). It includes fundus images showing 13 diseases associated with human eye. It provides the list of disease codes and names for each image. Blood vessels and optic nerve have the pixel level annotation but without grading. Two observers manually segmented all the images. On average, the first person labeled 32,200 pixels in each image as vessel, while the second person labeled 46,100 pixels in each image as vessel. This dataset offers a challenging OD detection problem due to the appearance of retinal diseases.

\subsubsection{DIARETDB1}

This dataset contains 89 color fundus images, which were taken under varying imaging settings and FOV of 50$^{\circ}$, and were captured in Kuopio university hospital in Finland \cite{kalviainen2007diaretdb1}. Four independent experts annotated the images. These experts delineated the regions where MAs and HMs can be found, and provided a map for each type of lesion. This dataset is referred to as “calibration level 1 fundus images”. It is divided into training and test sets containing 28 and 61 images, respectively.
 
\subsubsection{CHASE}
It was acquired under the program child heart and health study in England (CHASE) \cite{chase} from children of different ethnic origin and ages from 9 and 10 years. It consists of 28 fundus images taken from 14 children and the annotation contains ground truths for blood vessels collected using Top Con TRV-50 camera with 35 FOV. Unlike DRIVE and STARE, it contains images with uneven and non-uniform background illumination, poor contrast of blood vessels and wider arteries that have a bright strip running down the center, known as the central vessel reflex.
 
\subsubsection{DRiDB}
Diabetic retinopathy image database (DRiDB) \cite{prentasic2013diabetic} was obtained at university hospital in Zagreb and was created to overcome the shortcomings in previous datasets such as grading and limited number of observers. Images were taken and selected by experts with 45 FOV and shown DR symptoms vary from almost normal to cases where new fragile vessels are visible. In this dataset, each image was evaluated by five independent experts to mark DR findings. These experts annotated pixels of findings and related areas of MAs, HMs, hard and soft EXs, blood vessels, ODs and macula.

\subsubsection{ORIGA}

Online retinal fundus image database for glaucoma analysis and research (ORIGA) \cite{zhang2010origa} is an online repository which shares fundus images and their ground truths as benchmarks for researchers to share retinal image analysis results and the corresponding diagnosis. It was collected over a period of 3 years from 2004 to 2007 at Singapore eye research institute. It focuses on OD and optic cup (OC) segmentation and Cup-to-Disc Ratio (CDR) to diagnosis glaucoma. 
 
\subsubsection{SCES}
It was acquired under Singapore Chinese eye study (SCES) \cite{sng2012determinants} conducted on 1,060 Chinese participants and was graded by one senior professional grader and one retinal specialist. The study was conducted to identify the determinants of anterior chamber depth (ACD) and to ascertain the relative importance of these determinants in Chinese persons in Singapore.

\subsubsection{AREDS}

It was  developed under Age-related eye disease study (AREDS) \cite{nih}, which was funded by NIH. It is long-term multicenter, prospective study of 595 participants with ages from 55 to 80 years, which was designed to assess the clinical course of both AMD and cataract. Participants were of any illness or condition that would make long-term follow-up. On the basis of fundus photographs graded by a central reading center, best corrected visual acuity, and ophthalmologic evaluations, participants were enrolled in one of several AMD categories.

\subsubsection{REVIEW}
 Retinal vessel image set for estimation of widths (REVIEW) \cite{al2008reference} was made available online in 2008 by the department of computing and informatics at the university of Lincoln, UK. The dataset contains 16 mydriatic images with 193 annotated vessel segments consisting of 5066 profile points manually marked by three independent experts. Unlike DRIVE and STARE, REVIEW dataset includes width measurements. The images were chosen to evaluate the accuracy and precision of the vessel width measurement algorithms in the presence of pathology and central light reflex. The 16 images are subdivided into four sets: the high resolution image set (8 images), the vascular disease image set (4 images), the central light reflex image set (2 images) and the kickpoint image set (2 images). %no grading

\subsubsection{EyePACS-1}

 Eye picture archive and communication system (EyePACS) \cite{eyepacs} is a flexible protocol and web based telemedicine system for DR screening and collaboration among clinicians. Patients’ fundus images can be easily uploaded to EyePACS web. The protocol evaluates the presence and severity of discrete retinal lesions associated with DR. The protocol uses the Canon CR-DGi and Canon CR-1 nonmydriatic cameras can be accessed on the EyePACS Web site. The lesions are graded as MAs, HMs with or without MAs, cotton wool spots, intraretinal microvascular abnormalities, venous beading, new vessels (new vessels on the disk and new vessels elsewhere), fibrous proliferation, vitreous HMs or preretinal HMs and HEs. In addition, the presence or absence of laser scars. Graders grade each lesion type separately in each image using an online grading template that records a choice for each lesion type among no (absent), yes (present) or cannot grade.
 
\subsubsection{RIM-ONE}
It is an open retinal image database for optic nerve evaluation (RIM-ONE) \cite{fumero2011rim} captured by non-mydriatic Nidek AFC-210 with a body of a Canon EOS 5D Mark II. It was designed for glaucoma diagnosis and consists of 169 optic nerve head regions, which were cropped manually from full fundus images. These images were annotated by 5 glaucoma experts: 4 ophthalmologists and 1 optometrist.

\subsubsection{DRISHTI-GS} 
It consists of a total of 101 fundus images of healthy controls and glaucoma patients with almost 25 FOV that were collected at Aravind eye hospital in India \cite{sivaswamy2014drishti}. It is divided into training and test sets consisting of 50 and 51 images, respectively. All images were annotated by 4 ophthalmologists with clinical experiences of 3, 5, 9 and 20 years, respectively. The manual segmentation of OD and OC boundaries, and CDR are provided as ground truths. Also two other expert opinions were included about whether an image represents healthy control or glaucomatous eye and presence or absence of notching in the inferior and/or superior sectors of the image.
  
\subsubsection{ARIA}  
Automated retinal image analyzer (ARIA) \cite{aria} was collected and designed to trace blood vessels, ODs and fovea locations. It was marked by two image analysis experts. This dataset was collected at St Paul’s eye unit and the university of Liverpool to diagnosis AMD and DR using a Zeiss FF450+ fundus camera at a 50 FOV.
 \subsubsection{DRIONS-DB}
Digital retinal images for optic nerve segmentation database (DRIONS-DB) \cite{drion} was collected at a university hospital in Spain. It was designed to segment optic nerve head and its related pathologies. It was annotated by 2 independent medical experts. Images were centered on the optic nerve head and are stored in slide format.
  
  \subsubsection{SEED}
Singapore epidemiology of eye diseases (SEED) \cite{seed} was composed of 235 fundus images with a focus on studying major eye diseases, including DR, AMD, glaucoma, refractive errors and cataract. Each image has OD and OC regions marked by a trained grader, which serves as a ground truth for segmentation.

  \begin{table}[H]
\centering
	\caption{Datasets for DR Detection}
		\label{dstbl}
	\begin{tabular}{|l|l|l|l|l|}
		\hline
		\rowcolor[HTML]{EFEFEF} 
		Dataset                                                           & \#Images                                                                                       & Resolution                                                                     & Format &                                                                                                  Tasks                                                                                                              \\ \hline
	    %%%%
		\multicolumn{5}{|c|}{\cellcolor[HTML]{EFEFEF}Images level annotation} \\ \hline
		%%%%%
		MESSIDOR \cite{decenciere2014feedback}                                                 & 1,200                                                                                           & \begin{tabular}[c]{@{}l@{}}1,440$\times$960,\\   2,240$\times$1,488,\\2,304$\times$1,536\end{tabular} &\begin{tabular}[c]{@{}l@{}} Images: TIFF\\Diagnosis: excel file\end{tabular}                                                                                              & \begin{tabular}[c]{@{}l@{}}-DR grading\\   -Risk of DME\end{tabular}                                          \\ \hline
		%%%%%%%%%
			\begin{tabular}[c]{@{}l@{}}Kaggle\\ \cite{kgl}\end{tabular}        & 80,000                                                                                              & -                                                                              & JPEG      &\begin{tabular}[c]{@{}l@{}}-No DR \\-Mild\\ -Moderate \\ -Severe\\ -PDR\end{tabular}   \\ \hline                   
		%%%%%%%%%%
			AREDS\cite{nih}                                                            & 72,000                                                                                          & -                                                                              & -                                                         & -AMD  stages                                                                                                             \\ \hline
			%%%%%%%%%
				EyePACS-1\cite{eyepacs}                                                        & 9,963                                                                                          & -                                                                              & -     & \begin{tabular}[c]{@{}l@{}}-Referable DR\\ -MA\end{tabular}\\ \hline
			%%%%%%%%%%
				\multicolumn{5}{|c|}{\cellcolor[HTML]{EFEFEF}Pixel level annotation} \\ \hline
				%%%%%%%%%%%%%%%%%%%%%%%%%%%%
		
		\begin{tabular}[c]{@{}l@{}}e-ophtha\cite{optha}\end{tabular}      & \begin{tabular}[c]{@{}l@{}}148 MAs,\\   233 normal non-MA\\   47 EXs,\\ 35 normal non-EX\end{tabular} &\begin{tabular}[c]{@{}l@{}}2,544 $\times$ 1,696\\1440$\times$960\end{tabular} -                                                                              & 	\begin{tabular}[c]{@{}l@{}}Images: JPEG \\GT: PNG \end{tabular}                                                   & \begin{tabular}[c]{@{}l@{}}-MA\ small HM detection\\   -EX detection\end{tabular}                                     \\ \hline

%%%%%%%%%%%%%%%%%%%%%%%%%	                                                                                 
		DRIVE \cite{drive}                                                  & \begin{tabular}[c]{@{}l@{}}33 normal\\7 mild to early\\ DR stage\end{tabular}                  & 584$\times$565                                                                              &  \begin{tabular}[c]{@{}l@{}}Images: TIFF\\GT, masks: GIF\end{tabular}                                                & -Vessels extraction                                                                                               \\ \hline
		%%%%%%%%
		\begin{tabular}[c]{@{}l@{}}STARE\\  \cite{hoover2000locating}\end{tabular}        & 402                                                                                            & 605$\times$700                                                                        & PPM                                  & \begin{tabular}[c]{@{}l@{}}-13 retinal diseases\\ -Vessels extraction\\ -Optic nerve\end{tabular}                 \\ \hline
		%%%%%%%		
		\begin{tabular}[c]{@{}l@{}}DIARETDB1\\   \cite{kalviainen2007diaretdb1,dia}\end{tabular} & \begin{tabular}[c]{@{}l@{}}5 normal\\84 with at least one\\NPDR sign\end{tabular}              & 1,500$\times$1,152                                                                              & \begin{tabular}[c]{@{}l@{}}Images,\\ masks,\\ GT: PNG\end{tabular}&  \begin{tabular}[c]{@{}l@{}}-MAs\\-HMs\\-SEs\\-HEs\end{tabular}                                                                                                            \\ \hline
		%%%%%%%%
		CHASE  \cite{chase}                                                           & 28                                                                                     & 1,280 $\times$ 960                                                                                    &  \begin{tabular}[c]{@{}l@{}}Images: JPEG\\GT: PNG\end{tabular}                               & -Vessels extraction                                                                                               \\ \hline
		%%%%%%%%%%%%%%%%%%%%%
		DRiDB  \cite{prentasic2013diabetic}                                                           & 50                                                                                             & 720$\times$576                                                                             & BMP     & \begin{tabular}[c]{@{}l@{}}-MAs\\ -HMs\\ -HEs\\ -SEs\\ -Vessels extraction\\ -OD\\ -Macula\end{tabular} \\ \hline
		%%%%%%%%%%%%%%%%%%%
		ORIGA\cite{zhang2010origa}                                                          & \begin{tabular}[c]{@{}l@{}} 482 normal\\168 glaucomatous\end{tabular}                         & 720$\times$576                                                                              & -                                                          & \begin{tabular}[c]{@{}l@{}}-OD\\ -Optic cup\\ -Cup-to-Disc Ratio (CDR)\end{tabular}                               \\ \hline
	%%%%%%%%%%%%%%%%
		SCES\cite{sng2012determinants}                                                             & \begin{tabular}[c]{@{}l@{}}1,630 normal\\46 glaucomatous\end{tabular}                         & -                                                                              & -                                                    & -CDR                                                                                          \\ \hline
	%%%%%%%%
		REVIEW\cite{al2008reference}                                                           & 16                                                                                              &\begin{tabular}[c]{@{}l@{}}3,584$\times$2,438\\1,360$\times$1,024\\2,160$\times$1,440\\3,300$\times$2,600\end{tabular}                                                                                & -                                            & -Vessels extraction                                                                                               \\ \hline
		%%%%%%%%%%%%

		RIM-ONE\cite{fumero2011rim}&\begin{tabular}[c]{@{}l@{}}118 normal
		\\12 early glaucoma
		 \\14 moderate glaucoma
	\\  14 deep glaucoma
		\\ 11 ocular hypertension\end{tabular} &-&-&-Optic nerve
	\\ \hline
	%%%%%%%%%%%%
	DRISHTI-GS\cite{sivaswamy2014drishti}&\begin{tabular}[c]{@{}l@{}}31 normal\\70 glaucomatous\end{tabular} &2,896 $\times$ 1,944&PNG&\begin{tabular}[c]{@{}l@{}}-OD segmentation\\-OC segmentation\end{tabular} \\ \hline
	%%%%%%%%%%%%	
	ARIA \cite{aria} &\begin{tabular}[c]{@{}l@{}}16 normal\\92 AMD \\59 DR\end{tabular}&768$\times$576&TIFF&\begin{tabular}[c]{@{}l@{}}-OD\\-Fovea location\\-Vessel extraction\end{tabular}\\ \hline
%%%%%%%%%%%%%%%	
	DRIONS-DB\cite{drion}&110&600$\times$400&\begin{tabular}[c]{@{}l@{}}Images: JPEG\\GT: txt file\end{tabular}&-OD\\ \hline
	%%%%%%%%%%%%%%%%
	SEED-DB\cite{seed}&\begin{tabular}[c]{@{}l@{}}192 normal\\43 glaucomatous\end{tabular}&3,504$\times$ 2,336&-&\begin{tabular}[c]{@{}l@{}}-OD\\-OC\end{tabular}\\ \hline
	\end{tabular}
\end{table}

\subsection{Performance Metrics}

In this section, we define the performance metrics that are commonly used to assess DR detection algorithms. Common metrics for measuring the performance of classification algorithms include accuracy, sensitivity (recall), specificity, precision, F-score, ROC curve, logloss, IOU, overlapping error, boundary-based evaluation and the dice similarity coefficient.

Accuracy is defined as the ratio of the correctly classified instances over the total number of instances \cite{mitchell2004role}. It is formally defined as:

	\begin{equation}
	Accuracy=\frac{TP+TN}{TP+TN+FP+FN},
	\end{equation}
where $TP$ (true positive) is the number of positive instances (e.g., having DR) in the considered dataset that are correctly classified, $TN$ (true negative) is the number of negative instances (e.g., normal cases) in the considered dataset that are correctly classified, and $FP$ (false positive) and $FN$ (false negative) are the numbers of positive and negative instances that are incorrectly classified, respectively. Note that in detecting DR, an instance is either a fundus image, a patch or a pixel of a fundus image, depending on the task. $\textit{Sensitivity(SN)}$, or the true positive rate or recall, measures the fraction of correctly classified positive instances; $\textit{specificity(SP)}$, or the true negative rate, measures the fraction of correctly classified negative instances; and precision, or positive predictive value, measures the fraction of positive instances that are correctly classified. They are formally defined as follows:
	
	\begin{equation} \label{eq:sens}
	Sensitivity(Recall)=\frac{TP}{TP+FN}
	\end{equation} 
	\begin{equation}\label{eq:spec}
	Specificity=\frac{TN}{TN+FP}
	\end{equation}
	\begin{equation}\label{eq:pre}
	Precision=\frac{TP}{TP+FP}
	\end{equation}
	
$\textit{F-score}(F)$ combines precision and recall as follows:

	\begin{equation} \label{eq:fscore}
	F=2\frac{Precision\times Recall}{Precision+Recall}
	\end{equation} 	
The $\textit{receiver operating characteristic (ROC)}$ curve represents the plot of the $\textit{true positive rate}$ against the $\textit{false positive rate}$. It shows the relationship between sensitivity and specificity. The $\textit{area under the ROC curve (AUC)}$ is also used as a performance metric and takes values between 0 and 1; the closer the AUC is to 1, the better the performance. $\textit{Logarithmic loss (log loss)}$ determines a classifier’s accuracy by penalizing false classifications. To find log loss, the classifier must assign a probability to each class, instead of presenting the most likely class. It is given by:

	\begin{equation}
	logloss=-\frac{1}{N} \sum_{i=1}^{N} \sum_{j=1}^{M}y_{ij} log p_{ij},
	\end{equation}
	
	where N is the number of samples, $M$ is the number of labels, $y_{ij}$ is a binary indicator of whether label $j$ is the correct classification for instance $i$, and $p_{ij}$ is the model’s probability of assigning label $j$ to instance $i$.		As segmentation is also a kind of classification at the pixel level, the metrics defined for classification can be used for segmentation. Additional metrics used for measuring the performance of segmentation algorithms include overlapping error, intersection over union and the dice similarity coefficient. $\textit{Intersection over union (IOU)}$ is defined as follows \cite{shankaranarayana2017joint}:

	\begin{equation}
	IOU=\frac{Area(A\cap G)}{Area(A\cup G)}
	\end{equation}
 	$\textit{Overlapping error}$ is obtained by:
	\begin{equation}
	E=1-IOU
	\end{equation}

where $A$ is the notation for segmentation of the output and $G$ indicates the manual ground truth segmentation \cite{srivastava2015using}.

$\textit{Boundary-based evaluation (B)}$ is the absolute pointwise localization error obtained by measuring the distance between two closed boundary curves. Let $C_g$ be the boundary of ground truth and $C_a$ be the boundary obtained from a method. The distance $D$ between two curves is defined as (in pixels):

\begin{equation}
B=\frac{1}{n}\sum_{\theta=1}^{\theta_n}\sqrt{(d_{g}^{\theta})^2-(d_{a}^{\theta})^2},
\end{equation}

where $d_{g}^{\theta}$ and $d_{a}^{\theta}$ are the distance from the centroid of the curve to points on $C_g$ and $C_a$ in the direction of $\theta$ and $n$ is the total number of angular samples. The distance between the calculated boundary and ground truth should ideally be close to zero \cite{joshi2011optic}.
 
An alternative to overlapping error that is used for DR detection is the $\textit{dice similarity coefficient (DSC)}$ or overlap index, which is defined by \cite{zou2004statistical}:

		\begin{equation}
		DSC = \frac{2 TP}{2 TP+FP+FN}
		\end{equation}
	
The DSC takes values between 0 and 1; the closer the DSC is to 1, the better the segmentation results are. 	$\textit{Region precision recall (RPR)}$ is commonly used to assess edge or boundary detection outcomes based on region overlapping. It refers to the segmentation quality in a precision recall space \cite{zhang2015segmentation}.

	\section{Overview of Deep Learning}\label{Deep learning}

Various DL-based architectures have been introduced. Some commonly employed deep architectures for various DR-detection include convolutional neural networks (CNNs), autoencoders (AEs), recurrent neural networks (RNNs) and deep belief networks (DBNs). In the following paragraphs, we give an overview of these architectures.

\subsection{Convolutional Neural Networks}

CNNs simulate the human visual system and have been widely employed for various computer vision tasks. They mainly consist of three types of layers: convolutional, pooling and fully connected (FC). Convolutional layers employ a convolution operation to encode local spatial information and then FC layers to encode the global information. Commonly used CNN models include AlexNet, VGGNet, GoogLeNet and ResNet. A CNN model is taught in an end-to-end manner; it learns the hierarchy of features automatically and results in outstanding classification performance. Initial CNN models such as LeNet \cite{lecun1998gradient} and AlexNet \cite{krizhevsky2012imagenet} contain few layers. In 2014, \citet{simonyan2014very} explored a deeper CNN model called VGGNet, which contains 19 layers, and found that depth is crucial for better performance. Motivated by these findings, deeper models such as GoogLeNet, Inception \cite{szegedy2015going} and ResNet \cite{he2016deep} have been proposed, which have shown amazing performance in many computer vision tasks. An end-to-end model usually means a deep model that takes inputs and gives outputs. Transfer learning means that a model is first taught in an end-to-end fashion using a dataset from a related domain and then fine-tuned using the dataset from the domain. 
Learning a CNN model requires a very large amount of data to overcome overfitting problems and ensure proper convergence \cite{tajbakhsh2016convolutional}, but large amounts of data are not available in the medical domain, particularly for DR detection. The solution is to use transfer learning \cite{erhan2010does}. Generally, two strategies of transfer learning are used: (i) using a pre-trained CNN model as a feature extractor and (ii) fine-tuning a pre-trained CNN model using data from the relevant domain. A fully convolutional network (FCN) is a version of a CNN model in which FC layers are converted into convolutional layers and deconvolution (or transposed convolution) layers are added to undo the effect of down-sampling during the convolutional layers and to obtain an output map of the same size as the input image \cite{long2015fully}. This model is commonly used for segmentation.

\subsection{Autoencoder-based and Stacked Autoencoder Methods}
An autoencoder (AE) is a single hidden layer neural network with the same input and output \cite{hinton2006reducing} and is used to build a stacked-autoencoder (SAE), a deep architecture \cite{liou2014autoencoder}. The training of an SAE model consists of two phases: pre-training and fine-tuning. In the pre-training phase, an SAE is trained layer by layer in an unsupervised way. In the fine-tuning phase, the pre-trained SAE model is fine-tuned using gradient descent and backpropagation algorithms in a supervised way. An autoencoder is the basic building block of an SAE. There two main types of autoencoders: sparse and denoising. $\textit{Sparse autoencoders}$ are a type of autoencoder that tends to force the extracting of sparse features from raw data. The sparsity of the representation can either be achieved by penalizing hidden unit biases or by directly penalizing the output of hidden unit activations. $\textit{Denoising autoencoders (DAEs)}$ have also been used in DR detection \citet{maji2015deep} due its robustness in recovering the corrupted input and force the model to capture the correct version.

\subsection{Recurrent Neural Networks}
RNNs are a type of neural network that learns the context as well along with input patterns. It learns the output of the previous iterations and combines it with the current input to yield an output; in this way, an RNN is able to influence itself through recurrences. An RNN model usually contains three sets of parameters—input to hidden weights $W$, hidden weights $U$ and hidden weights—to output$V$ where weights are shared across position/time of input sequence \cite{mikolov2010recurrent}.
	
\subsection{Deep Belief Networks}
A DBN \cite{vinyals2017show} is a deep network architecture that is built with cascading restricted Boltzmann machines (RBMs). An RBM is taught using a contrastive divergence algorithm in such a way that maximizes the similarity (in the sense of probability) between the input and its projection. The involvement of probability as a similarity measure prevents degenerate solutions and makes DBNs a probabilistic model. Just like SAEs, DBNs are first pre-trained in an unsupervised way using a layer-by-layer greedy learning strategy; then, it is fine-tuned using gradient descent and backpropagation algorithms.

\section{Literature Survey}\label{LS}
Based on the clinical importance of DR detection tasks, we categorize them into four categories: (i) retinal blood vessel segmentation, (ii) optic disk localization and segmentation, (iii) lesion detection and classification, and (iv) image-level DR diagnosis for referral. In the following sub-sections, we review the state-of-the-art DL-based algorithms for these tasks.	

\subsection{Retinal Blood Vessel Segmentation}
It is very important to identify changes in fine retinal blood vessels for preventing vision impairment due to pathological retinal damage. The segmentation of retinal blood vessels is challenging due to their low contrast, variations in their morphology against a noisy background and the presence of pathologies like MAs and HMs. Different learning approaches have been applied to segment retinal blood vessels. In the following paragraphs, we review these methods based on DL approaches.

\subsubsection{Convolutional Neural Networks} 

 Many retinal blood vessel segmentation algorithms based on CNN models have been proposed. \citet{maji2016ensemble} employed 12 CNN models to segment vessel and non-vessel pixels. Each CNN model consists of three convolutional layers and two fully connected layers. For evaluation, they used the DRIVE dataset.

\citet{liskowski2016segmenting} proposed a pixel-wise supervised vessel-segmentation method based on deep CNN, which is trained using fundus images that have been pre-processed with global contrast normalization and zero-phase whitening, and augmented using geometric transformations and gamma corrections. They used the DRIVE, STARE and CHASE datasets to evaluate the system. It is robust against the central vessel reflex and sensitive in detecting fine vessels.

\citet{maninis2016deep} formulated the retinal blood vessel segmentation problem as an image-to-image regression task, for which they employed pre-trained VGG, which was modified by removing FC layers and incorporating additional convolutional layers after the first four convolution blocks of VGG before pooling the layers. The additional convolutional layers are upsampled to the same size as the image, trained and concatenated into a volume. They used DRIVE and STARE for evaluation.

\citet{wu2016deep} first extracted discriminative features using a CNN and then used nearest neighbor search based on principal component analysis (PCA) to estimate the local structure distribution, which was finally employed by the generalized probabilistic tracking framework to segment blood vessels. This method was evaluated using the DRIVE dataset. 

\citet{dasgupta2017fully} used an FCN combined with structured prediction to segment blood vessels, which they assumed to be a multi-label inference task. The green channel of the images was preprocessed by normalization, contrast, gamma adjustment and scaling the intensity value between 0 and 1. They used DRIVE to evaluate the method’s performance. 

\citet{tan2017segmentation} proposed a seven-layer CNN model to simultaneously segment blood vessels, OD and fovea. After normalizing the colored images, they formulated the segmentation problem as a classification problem assuming four classes—blood vessels, OD, fovea and background—and classified each pixel by taking a neighborhood of 25$\times$25 pixels. This is very time consuming because each pixel is classified independently, with as many passes made through the net as the number of pixels. Its performance was evaluated with the DRIVE dataset.

 \citet{fu2016retinal} similarly formulated the blood vessel segmentation problem as a boundary-detection task and proposed a method for this task by integrating FCN and fully connected conditional random field (CRF). First, a vessel probability map is created using FCN, and then the vessels are segmented by combining the vessel probability map and long-range interactions between pixels using CRF. This method was validated on the DRIVE and STARE datasets. 
 
\citet{mo2017multi} used an FCN and incorporated some auxiliary classifiers in intermediate layers to make the features more discriminative in lower layers. To overcome the small number of available samples, they used transfer learning to train the FCN model. They evaluated the system on DRIVE, STARE and CHASE.

The performance analysis of all the aforementioned methods is given in Table \ref{vessels}. This analysis indicates that among all CNN-based methods, the one by \citet{liskowski2016segmenting} performed better than all other methods in terms of accuracy, AUC and sensitivity. This method may outperform due to the preprocessing of fundus images and training of the CNN model using an augmented dataset. All of the other methods use pre-trained CNN models without preprocessing or augmentation. Against expectations, the ensemble of CNN models by \citet{maji2016ensemble} did not perform better than the other CNN-based methods because there is no preprocessing and augmentation of the training dataset.

\subsubsection{Stacked Autoencoder-Based Methods} 
Some methods employ SAEs in different ways to segment vessels. The method proposed by \citet{maji2015deep} uses a hybrid DL architecture, which consists of unsupervised stacked denoising autoencoders (SDAEs), to segment vessels in fundus images. The structure of the first DAE consists of 400 hidden neurons, and the second DAE contains 100 hidden neurons. SDAE learns features, which are classified using random forest (RF). This approach segments vessels using patches of size k × k around each pixel in the green channel. They used DRIVE to assess the method.

 \citet{roy2015dasa} introduced an SAE-based deep neural network (SAE-DNN) model for vessel segmentation that employs the domain adaptation (DA) approach for its training. SAE-DNN consists of two hidden layers, which are trained using the source domain (DRIVE dataset), using an auto-encoding mechanism and supervised learning. Then, DA is applied in two stages: unsupervised weight adaptation and supervised fine-tuning. In unsupervised weight adaptation, hidden nodes of the SAE-DNN are re-trained using unlabeled samples from the target domain (STARE dataset) with the auto-encoding mechanism using systematic node dropouts, whereas in supervised fine-tuning, the SAE-DNN is fine-tuned using a small number of labeled samples from the target domain. The results show that domain DA improves the performance of the SAE-DNN.
 
 \citet{li2016cross} proposed segmenting retinal vessels from the green channel using a supervised DL approach that labels the patch of a pixel instead of a single pixel. In this approach, the vessel-segmentation problem is modeled as a cross-modality data transformation that transforms a retinal image to a vessel map and is defined using a deep neural network consisting of DAEs. They assessed the performance on DRIVE, STARE and CHASE (28 images).

 \citet{lahiri2016deep} used a two-level ensemble of stacked denoised autoencoder networks (SDAEs). In the first-level ensemble, a network (E-net) consists of $n$ SDAEs composed of the same structure; each SDAE contains two hidden layers and is followed by a Softmax classifier; SDAEs are trained on bootstrap training samples using an auto-encoding mechanism in parallel, to produce probabilistic image maps, which are conglomerated using a fusion strategy. In the second level of the ensemble, to introduce further diversity, decisions from two E-nets having different architectures are merged using the convex weighted average. The authors used the DRIVE dataset to evaluate the method.

\subsubsection{Recurrent Neural Network-Based Methods}
\citet{fu2016deepvessel} formulated the blood vessel segmentation problem as a boundary detection task and proposed the DeepVessel method by integrating CNN and CRF as an RNN and evaluated it on the DRIVE, STARE, and CHASE datasets.

A performance analysis of the aforementioned methods is given in Table \ref{vessels}. This analysis indicates that among all SAE-based methods, the methods based on cross-modality transformation \cite{li2016cross} and two-level ensemble of SAEs \cite{lahiri2016deep} outperform SAE-based methods in terms of accuracy. Although the method based on two-level ensemble of SAEs \cite{lahiri2016deep} performed slightly better than the method based on cross-modality transformation \cite{li2016cross}, the difference was not significant. Interestingly, there was no noticeable difference in the performance of methods based on CNNs and on SAEs in terms of accuracy. CNN models involve much more learnable parameters than SAE models and as such are prone to overfitting. CNN models can do better provided a huge labeled dataset is available or novel augmentation techniques are introduced.

\begin{center}
	
	\begin{table}[]
		\centering
		\caption{Representative of works in diabetic retinopathy (DR) vessels detection} 
		\label{vessels}
		
		\hspace*{-6cm} 
	\begin{tabular}{|l|p{3.5cm}|l|l|l|}
			\hline
		Research study & Segmentation Method                                                                                                                                                                     &Training& Dataset& Performance \\ \hline
		
			\multicolumn{5}{|c|}{\cellcolor[HTML]{EFEFEF}CNN-Based Methods} \\ \hline
		
			\citet{maji2016ensemble} & Patch-based ensemble of CNN models&End-to-end                                                                                                                                                                       & DRIVE & \begin{tabular}[c]{@{}l@{}} AUC=0.9283\\ACC= 94.7 \end{tabular} \\ \hline
				%%%%%%%%%%%%%%%%%%%%%%%
			
						\multirow{3}{*}{\citet{liskowski2016segmenting}}   & 	\multirow{3}{*}{Patch-based CNN}  &	\multirow{3}{*}{End-to-end}                                                               & DRIVE  &\begin{tabular}[c]{@{}l@{}}SN= 98.07, SP=78.11\\AUC=0.9790\\ACC=95.35\end{tabular} \\ \cline{4-5}

						&    &                                      &STARE                                                                                            
						& \begin{tabular}[c]{@{}l@{}}SN=85.54, SP=98.62 \\AUC=0.9928 \\ACC=97.29 \end{tabular}\\ \cline{4-5}
						&     &   &                                                          CHASE &\begin{tabular}[c]{@{}l@{}}SN=81.54, SP=98.66\\AUC=0.988\\ACC=96.96\end{tabular} \\ \hline
						
						%%%%%%%%%%%%%%%%%%%%%%%%%%%%%%%%
					
						\multirow{2}{*}{\citet{maninis2016deep}} 
						 &\multirow{2}{*}{ FCN}&\multirow{2}{*}{Transfer learning}   & DRIVE & RPR=0.822 \\ \cline{4-5} 
						&  &                                                                                                                                                                                                                                         & STARE&RPR=0.831    \\ \hline
						
						%%%%%%%%%%%%%%%%%%%%%%%%%%%%%%%%%%%%% 
						\citet{wu2016deep}                                     & \begin{tabular}[c]{@{}p{2.5cm}@{}}
					Vessel tracking/patch-based CNN/PCA as classifier\end{tabular}  &End-to-end                                                                 & DRIVE & AUC=0.9701\\ \hline
								
						%%%%%%%%%%%%%%%%%%%%%%%%%%%%%%%%%%%%%%%%
							%to classify each and every pixel contained
							\citet{dasgupta2017fully}    & Patch-based FCN&End-to-end                                                                   &DRIVE & \begin{tabular}[c]{@{}l@{}}SN=76.91\\SP=98.01  \\AUC=0.974\\ACC=95.33\end{tabular} \\ \hline						
							%%%%%%%%%%%%%%%%%%%%%%
								\citet{tan2017segmentation} & Patch-based seven-layers CNN
								 & End-to-end          & DRIVE& \begin{tabular}[c]{@{}l@{}}SN=75.37\\SP=96.94\end{tabular}\\ \hline	
								%%%%%%%%%%%%%%%%%%%%%%%%%%

					\multirow{2}{*}{\citet{fu2016retinal}}                    &  	\multirow{2}{*}{FCN/CRF} &	\multirow{2}{*}{End-to-end }                                                                                                                                                                    & DRIVE  &\begin{tabular}[c]{@{}l@{}}SN=72.94, ACC=94.70 \end{tabular} \\ \cline{4-5} 
					&    &                                                                                                                                                                      & STARE  &\begin{tabular}[c]{@{}l@{}}SN=71.40, ACC=95.45\end{tabular}\\ \hline
					%%%%%%%%%%%%%%%%%%%%%%%%%%%%%%

					\multirow{3}{*}{\citet{mo2017multi}}&\multirow{3}{*}{Multi-level FCN} &\multirow{3}{*}{End-to-end}                                                                                                  & DRIVE &\begin{tabular}[c]{@{}l@{}}SN=77.79, SP=97.80\\AUC=0.9782\\ACC=95.21\end{tabular}\\ \cline{4-5} 
					& &                                                                                                   & STARE   & \begin{tabular}[c]{@{}l@{}}SN=81.47, SP=98.44\\AUC=0.9885\\ACC=96.76\end{tabular} \\ \cline{4-5} 
					&		
					 &					                                                                                                     & CHASE &\begin{tabular}[c]{@{}l@{}}SN=76.61,  SP=98.16\\AUC=0.9812\\ACC=95.99\end{tabular}                                                                             \\ \hline
					
					%%%%%%%%%%%%%%%%%%%%%%%%%%%%%%%

	\multicolumn{5}{|c|}{\cellcolor[HTML]{EFEFEF}AE-Based Methods} \\ \hline
			%%%%%%%%%%%%%%%%%%%%%%%%%%%%%%%%%%
		%	\multicolumn{8}{|c|}{\cellcolor[HTML]{FFEFFF}Pixel Level Methods} \\ \hline
				\citet{maji2015deep}                                      & Patch-based SDAE/RF  &Transfer learning                                                                 & DRIVE  &\begin{tabular}[c]{@{}l@{}}AUC=0.9195, ACC=93.27\end{tabular}\\ \hline
				%%%%%%%%%%%%%%%%%%%%
					\citet{roy2015dasa}  & Patch-based SAE&Transfer learning  &STARE &\begin{tabular}[c]{@{}l@{}}AUC=0.92 \\ logloss=0.18\end{tabular} \\ \hline
					% & SAE&Transfer learning
					%&  STARE&  - & -&0.87      & logloss=0.39\\ \hline
					%%ROC=area under ROC written
					%%%%%%%%%%%%%%%%%%%%%%%%%%%%%%%%%%%%%%%%%%
			\multirow{3}{*}{\citet{li2016cross}}                      & \multirow{3}{*}{Patch-based SDAE} &\multirow{3}{*}{End-to-end}                                                                                                                                                                     & DRIVE &\begin{tabular}[c]{@{}l@{}}SN=75.6, SP=98\\AUC=0.9738\\ACC=95.27\end{tabular}\\ \cline{4-5} 
			&    &                                                                                                                                                                      & STARE&\begin{tabular}[c]{@{}l@{}}SN=77.26, SP=98.79\\ACC=96.28\end{tabular}                                                                \\ \cline{4-5} 
			&    &                                                                                                                                                                    & CHASE (28 images)&\begin{tabular}[c]{@{}l@{}}SN=75.07, SP=97.93\\AUC=0.9716\\ACC=95.81\end{tabular}                                                              \\ \hline
			
			%%%%%%%%%%%%%%%%%%%%%%%%%%%%%%%%%%
	
			\citet{lahiri2016deep}    &Patch-based DSAE &Transfer learning                              &DRIVE&ACC=95.3 \\ \hline
		
				%%%%%%%%%%%%%%%%%%%%%
					\multicolumn{5}{|c|}{\cellcolor[HTML]{EFEFEF}RNN-Based Methods} \\ \hline
					\multirow{3}{*}{\citet{fu2016deepvessel}} & \multirow{3}{*}{\begin{tabular}[c]{@{}l@{}}Patch-based CNN/CRF\\as RNN\end{tabular}}  &	\multirow{3}{*}{End-to-end}&DRIVE 
					
					 &\begin{tabular}[c]{@{}l@{}}SP=76.03\\ACC=95.23 \end{tabular}\\ \cline{4-5} 
					&  &                                                                                                                                                                                & STARE  &\begin{tabular}[c]{@{}l@{}} SP=74.12 \\ACC=95.85\end{tabular} \\ \cline{4-5}
					&
					&                                                                                                                                                                                    
					& CHASE  &\begin{tabular}[c]{@{}l@{}}SP=71.30 \\ACC=94.89\end{tabular}  \\ \hline
					%%%%%%%%%%%%%%%%%%%%%%%%%%
				
		\end{tabular} \hspace*{-6cm}
	\end{table}

\end{center}
%%%%%%%%%%%%%%%%%%%%%%%%%%%%%%

\subsection{Optic Disc Feature}	
Detecting the OD can enhance DR detection and classification because its bright appearance can create confusion for other bright lesions such as EXs. OD detection involves two operations: (i) localizing and (ii) segmenting the OD. Both CNN and SAE models have been employed for OD detection. Some methods only localize the OD, which is basically an object-detection problem, and others localize and segment the OD, which is a segmentation problem, to identify the area of the OD along with its boundaries.
\subsubsection{Convolutional Neural Networks} 

\citet{lim2015integrated} method was one of the earliest proposals to employ a nine-layer CNN model to segment the OD and OC. It involves four main phases: localizing the region around the OD, enhancing this region by exaggerating the relevant visual features, classifying the enhanced region at pixel-level using a CNN model to produce a probability map and finally segmenting this map to predict the disc and cup boundaries. It was assessed on MESSIDOR and SEED-DB.

\citet{guo2016optic} used a large pixel patch-based CNN in which the OC was segmented by classification of each pixel patch and postprocessing. They used the DRISHTI-GS dataset for training and testing. 
Similarly, \citet{tan2017segmentation} segmented the OD and vessels jointly; it has been reviewed in section vessel segmentation.
\citet{sevastopolsky2017optic} used the modified U-net convolutional network presented in \cite{ronneberger2015u} to segment both the OD and OC.

 \citet{zilly2015boosting} ] introduced an OD and OC segmentation method based on a multi-scale two-layer CNN model that is trained with boosting. First, the region around the OD is cropped, down-sampled by a factor of 4, converted to L*a*b color space and normalized. Then, the region is processed by entropy filtering to identify the most discriminative points and is passed to the CNN model, which is trained using the gentle AdaBoost method. The logistic regression classifier produces a probability map from the output of the CNN model, and finally the graph cut method and convex-hull fitting are applied to get the segmented OD and OC regions. This method was evaluated with the DRISHTI-GS dataset using three performance metrics: F-score, overlap measure (IOU) and boundary error (B). 
 
 An extended version of this method is presented in \cite{zilly2017glaucoma}, \citet{zilly2017glaucoma} used ensemble CNN with entropy sampling to select informative points. These points were used to create a novel learning approach for convolutional filters based on boosting.

\citet{maninis2016deep} used the same FCN to segment both blood vessels and the OD from retinal, images as mentioned in the Vessel Segmentation section. The method was validated for OD and OC segmentation on the DRIONS-DB and RIM-ONE datasets. 

\citet{shankaranarayana2017joint} proposed a method for joint segmentation of the OC and OD using residual learning-based, fully convolutional networks (ResU-Net) that is similar to U-net \cite{ronneberger2015u} , which contains an encoder on the left side of the net involving down-sampling operations and a decoder on the right side employing up-sampling operations. A mapping between the retinal image and its segmentation map for OD and OC detection is trained using ResU-Net and generative adversarial networks (GANs). This method (ResU-GAN) does not involve any preprocessing and is efficient, compared with other pixel-segmentation methods \cite{lim2015integrated}. This method was tested using 159 images from the RIM-ONE dataset.

\citet{zhang2018automatic} used a faster region convolutional neural network (faster RCNN) with ZF net as the base CNN model to localize the OD. After localizing the OD, blood vessels in its bounding box are removed by using a Hessian matrix, and a shape-constrained level set is used to cut the OD’s boundaries. They used 4,000 images selected from Kaggle to train the CNN model and MESSIDOR for testing. This method is fast and gives very good localization results.

\citet{fu2018joint} used a U-shape CNN model (M-Net) to simultaneously segment the OD and OC in one stage and find the cup-to-disc ratio (CDR). The input layer of M-Net is a multi-scale layer consisting of an image pyramid. It involves a U-shaped CNN with a side-output layer to produce a local prediction map for different scale layers and a multi-label loss function output layer. First, the OD region is localized and transformed into a polar domain; then, it is passed through M-Net to generate a multi-label map, which is inverse transformed into the Cartesian domain to segment the OD. The ORIGA and SCES datasets were used to assess the method, which gave state-of-the-art results.

The method by \citet{niu2017automatic} used saliency map region proposal generation and a seven-layer-based CNN model to detect the OD. Using the saliency-based visual attention model, salient regions of a fundus image are identified, and a CNN model is used to classify these regions to locate the OD. This method is a validated DL approach that used cascading localization with feedback to localize the OD on preprocessed images using mean subtraction. The algorithm ends only when it finds a region containing the OD. The authors tested the performance on ORIGA,  MESSIDOR and these datasets together. 

\citet{alghamdi2016automatic} proposed a method for detecting abnormal ODs using a cascade of CNN models. First, candidate OD regions are extracted, preprocessed and normalized using whitening. Then, these regions are classified using the first module as the OD or non-OD. Finally, the detected OD regions are classified as normal, suspicious or abnormal by the second CNN module. This method was evaluated on DRIVE, DIARETDB1, MESSIDOR, STARE and a local dataset. 

\citet{xu2017optic} employed a pre-trained VGG model without the last FC layers and deconvolution layers connected to the last three pooling layers of a VGG model to calculate the probability map of pixels. The probability map is thresholded, and finally, the center of gravity of the pixels above the threshold is obtained to locate the OD. The authors used the ORIGA, MESSIDOR and STARE datasets for evaluation. This method is efficient in correctly localizing the OD.

Table \ref{OD} presents an aggregated view of the OD segmentation and localization methods. For OD segmentation, it is difficult to determine which method gives the best performance because all of the methods were evaluated on different databases using different metrics. Among the OD-localization methods, the method by \citet{zhang2018automatic} based on faster RCNN gives the best localization results for the MESSIDOR dataset.

\subsubsection{Stacked Autoencoder-Based Methods } 
We found just one method based on SAEs used to segment OD. \citet{srivastava2015using}'s idea is to distinguish parapapillary atrophy (PPA) from OD. This method crops the region of interest (ROI) after detecting the OD’s center and enhances its contrast using CLAHE. Features of each pixel are computed assuming a window of size 25x25 around it, which are then passed to a deep SAE consisting of one input layer with 626 units; seven hidden layers with 500, 400, 300, 200, 100, 50, and 20 units; and an output layer, to classify it as an OD or non-OD pixel. The binary map of the ROI obtained using the SAE is further refined for OD segmentation using an active shape model (ASM). The least mean overlapping error (LMOE) was used for evaluation on the dataset containing 230 images taken from ref. \cite{foong2007rationale}..
Table \ref{OD} provides a general view showing that CNN-based methods performed better than SAE-based methods.

\begin{table}[]
%	\centering
			\caption{OD detection works}
				\label{OD}	
	\centering	
	\hspace*{-6.5cm}
		\begin{tabular}{|p{3.5cm}|p{4.12cm}|p{2.5cm}|p{2.5cm}|p{2.1cm}|p{3.5cm}|}
	
		\hline

	Research study& Method&Training &Type(s) & Dataset  &Performance \\ \hline
		
\multicolumn{6}{|c|}{\cellcolor[HTML]{EFEFEF}CNN-Based Methods} \\ \hline
%%%%%%%%%%%%%%%%%%%%%%%
		%	\multicolumn{6}{|c|}{\cellcolor[HTML]{FFEFFF}Pixel Level Methods} \\ \hline
	\multirow{2}{*}{\citet{lim2015integrated}} & \multirow{2}{*}{\begin{tabular}[c]{@{}p{3cm}@{}}Nine-layer	CNN with exaggeration\end{tabular}} & \multirow{2}{*}{End-to-end} & \multirow{2}{*}{OD segmentation} & MESSIDOR & E=0.112, IOU=0.888\\ \cline{5-6} %;  S = 88.8
	& & & & SEED-DB  & E= 0.0843, IOU=0.916 \\ \hline %;  S = 91.6
	
	%%%%%%%%%%%%%%%%%%%%%%%%%%%%%%%%%%%%%%%%%%%%%%%%%
		\citet{guo2016optic}&Large pixel patch-based CNN&End-to-end&OC segmentation&DRISHTI-GS&F=93.73, E=0.1225\\ \hline
		%%%%%%%%%%%%%%%%%%%%%%%%%%%%%%%	
			\citet{tan2017segmentation} &Seven-layers CNN& End-to-end& \begin{tabular}[c]{@{}l@{}}OD segmentation\end{tabular} &DRIVE& ACC=87.90\\ \hline
			%ACC (pixels) = 87.90 \\ \hline
			
			%%%%%%%%%%%%%%%%%%%%%%%%%%%
				\multirow{4}{*}{\citet{sevastopolsky2017optic}} & \multirow{4}{*}{Modified U-Net CNN} & \multirow{4}{*}{Transfer learning} & \multirow{2}{*}{OD segmentation} & DRION-DB   & IOU=0.98, Dice=0.94 \\ \cline{5-6} 
				&                                     &                                    &                                  & RIM-ONE & IOU=0.98, Dice=0.95 \\ \cline{4-6} 
				&                                     &                                    & \multirow{2}{*}{OC segmentation} & DRION-DB   & IOU=0.75, Dice=0.85 \\ \cline{5-6} 
				&                                     &                                    &                                  & RIM-ONE & IOU=0.69, Dice=0.82 \\ \hline
				
				%%%%%%%%%%%%%%%%%%%%%%%%%%%%%%%%

	\multirow{2}{*}{\citet{zilly2015boosting}}& \multirow{2}{*}{\begin{tabular}[c]{@{}l@{}}Multi-scale two-layers CNN\end{tabular}} & \multirow{2}{*}{End-to-end} & OD segmentation & \multirow{2}{*}{DRISHTI-GS} & F=94.7, IOU=0.895, B=9.1 \\ \cline{4-4} \cline{6-6} 
	&                                                                                       &                             & OC segmentation &                             & F=83, IOU=0.864, B=16.5   \\ \hline
	
	%%%%%%%%%%%%%%%%%%%%%%%%%%%%%%%%%%%%%%%%%%%%%%%%

	\multirow{2}{*}{\citet{zilly2017glaucoma}} & \multirow{2}{*}{\begin{tabular}[c]{@{}l@{}}Ensemble learning-based\\ CNN\end{tabular}} & \multirow{2}{*}{End-to-end} & OD segmentation & \multirow{2}{*}{DRISHTI-GS} & F=97.3, IOU=0.914, B=9.9\\ \cline{4-4} \cline{6-6} 
	&                                                                                       &                             & OC segmentation &                             &F=87.1, IOU=0.85, B=10.2  \\ \hline
	%%%%%%%%%%%%%%%%%%%%%%%%%%			

		\multirow{2}{*}{\citet{maninis2016deep}} & \multirow{2}{*}{FCN based on VGG-16} & \multirow{2}{*}{Transfer learning} & \multirow{2}{*}{OD segmentation} & DRIONS-DB & RPR=0.971 \\ \cline{5-6} 
		&                                                 &                             &                                  & RIM-ONE  & RPR=0.959 \\ \hline
		
		%%%%%%%%%%%%%%%%%%%%%%% 
				\multirow{2}{*}{\begin{tabular}[c]{@{}p{2.7cm}@{}}\citet{shankaranarayana2017joint} \end{tabular}} & \multirow{2}{*}{\begin{tabular}[c]{@{}l@{}}ResU-Net and GANs\end{tabular}} & \multirow{2}{*}{Transfer learning} & OD segmentation & \multirow{2}{*}{RIM-ONE} & F=98.7, IOU= 0.961 \\ \cline{4-4} \cline{6-6} 
				&                                                                                       &                             & OC segmentation &                             &F=90.6, IOU=0.739  \\ \hline

				%%%%%%%%%%%%%%%%%%
					\multirow{2}{*}{\citet{zhang2018automatic}} & \multirow{2}{*}{Faster RCNN} & \multirow{2}{*}{Transfer learning} & OD localization & MESSIDOR                                                         & \begin{tabular}[c]{@{}l@{}}Mean average\\ precision=99.9\end{tabular}     \\ \cline{4-6} 
					&                              &                                    & OD segmentation & \begin{tabular}[c]{@{}l@{}}MESSIDOR \\ (120 images)\end{tabular} & \begin{tabular}[c]{@{}l@{}}Average matching \\ score of 85.4\end{tabular} \\ \hline		
					%%%%%%%%%%%%%%%%
					\citet{fu2018joint} &U-shaped CNN and polar transformation &Transfer learning& OD segmentation & ORIGA &E=0.071, IOU=0.929\\ \hline

					%%%%%%%%%%%%%%%%%%%%%%%%%%%%%%%%%
						\multirow{3}{*}{\citet{niu2017automatic}} & \multirow{3}{*}{\begin{tabular}[c]{@{}l@{}}Saliency map, CNN based \\ on AlexNet\end{tabular}} & \multirow{3}{*}{Transfer learning} & \multirow{3}{*}{OD localization} & ORIGA                                                       & ACC=99.33 \\ \cline{5-6} 
						&                                                                                                   &                                    &                                  & MESSIDOR                                                    & ACC=98.75\\ \cline{5-6} 
						&                                                                                                   &                                    &                                  & \begin{tabular}[c]{@{}l@{}}ORIGA+\\   MESSIDOR\end{tabular} & ACC=99.04 \\ \hline
						
						%%%%%%%%%%%%%%%%%%%	
					\multirow{4}{*}{\citet{alghamdi2016automatic}} & \multirow{4}{*}{\begin{tabular}[c]{@{}l@{}}Cascade CNN, each model \\ with 10-layers\end{tabular}} & \multirow{4}{*}{End-to-end} & \multirow{4}{*}{OD localization} & DRIVE     & ACC=100 \\ \cline{5-6} 
					&                                                                                                       &                             &                                  & DIARETDB1 & ACC=98.88 \\ \cline{5-6} 
					&                                                                                                       &                             &                                  & MESSIDOR  & ACC=99.20 \\ \cline{5-6} 
					&                                                                                                       &                             &                                  & STARE     & ACC=86.71 \\ \hline
					
					%%%%%%%%%%%%%%%%%%%%%%
				
					\multirow{3}{*}{\citet{xu2017optic}} & \multirow{3}{*}{\begin{tabular}[c]{@{}l@{}}CNN based on VGG and \\ deconvolution\end{tabular}} & \multirow{3}{*}{\begin{tabular}[c]{@{}l@{}}Transfer  learning\end{tabular}} & \multirow{3}{*}{OD localization} & ORIGA    & ACC=100   \\ \cline{5-6} 
					&                                                                                                   &                                                                &  & MESSIDOR & ACC=99.43 \\ \cline{5-6} 
					&                                                                                                   &                                                                                &                                  & STARE    & ACC=89 \\ \hline

					%%%%%%%%%%%%%%%%%%%%%%%%%%%%%

					\multicolumn{6}{|c|}{\cellcolor[HTML]{EFEFEF}AE-Based Methods} \\ \hline

				\citet{srivastava2015using} & SAE with ASM&End-to-end     & OD segmentation    & Local dataset used by \citet{foong2007rationale} &E=0.097  \\ \hline

				\end{tabular} \hspace*{-6.5cm}
				
			\end{table}
			%%%%%%%%%%%%%%%%%

\subsection{Lesion Detection and Classification}
Many DL methods have been proposed for detecting and classifying different types of DR lesions such as macular edema, exudates, microaneurysms and hemorrhages. In this section, we review these methods.

\subsubsection{Macula Edema as a Clinical Feature}	
The macula is the central part of retina, which consists of a thin layer of cells and light-sensitive nerve fibers at the back of eye, and is responsible for clear vision. Diabetic macula edema (DME) is a DR complication that occurs when the retinal capillaries become permeable and leakage occurs around the macula \cite{dr}; when vessels’ fluid and blood enter the retina, the macula swells and thickens. The DL methods for DME mainly can be categorized as CNN-based and AE-based methods.

\paragraph{Convolutional Neural Networks} 
 \citet{abramoff2016improved} proposed a supervised end-to-end CNN-based method to recognize DME. \citet{perdomo2016novel} proposed a method that combines EX localization and segmentation with DME detection. EX localization consists of two stages. In the first stage, an eight-layer CNN model, which takes a 488$\times$48 patch as its input, is used to localize EXs. It is trained on e-ophtha. In the second stage, using this CNN model as a predictor as well as the MESSIDOR dataset, grayscale mask images are produced. The DME detection model is based on the AlexNet architecture, which takes a fundus image together with a corresponding grayscale mask image as the inputs and predicts the class as normal, mild, moderate or severe DME. Preprocessing is used to extract the EXs’ ROIs, and data augmentation is applied to generate more samples to train the CNN model. The authors used MESSIDOR for testing.

\citet{burlina2016detection} used a deep convolutional neural network for feature extraction and a linear support vector machine (LSVM) for classification, for age-related macular degeneration (AMD). After cropping and resizing a fundus image to 231×231 pixels, the OverFeat CNN model pre-trained on the ImageNet dataset is used for feature extraction. The dataset NIH AREDS \cite{nih}, which is divided into four categories according to AMD severity, was used for validation.

 \citet{al2016diabetic} proposed an end-to-end CNN model for grading DME severity. After cropping and resizing a fundus image, red, green and blue channels are scaled to have zero mean and unit variance. The proposed CNN model consists of three convolution blocks and one block of FC layers. Data augmentation is applied to increase the number of samples for training. The model was evaluated using the MESSIDOR dataset.  
 
 \citet{ting2017development} evaluated the performance of a CNN model to diagnose AMD and other DR complications and concluded that their CNN was are effective in diagnosing DR complications but cannot identify all DME cases using fundus images. The CNN model for AMD detection was trained using 72,610 fundus images and was tested on 35,948 images from different ethnicities.

\citet{mo2018exudate} proposed a two-stage method to classify DME. In the first stage, a cascaded fully convolutional residual network (FCRN) with fused multi-level hierarchical information is used to create a probability map and segment EXs. In the second stage, using the segmented regions, the pixels with maximum probability are cropped and fed into another residual network to classify DME. They used the HEI-MED \cite{hei-med} and e-ophtha datasets to assess the method. 
 
 \paragraph{Deep Belief Networks} have also been employed for image-level DME diagnosis. 
 \citet{arunkumar2017multi} used a DBN for feature extraction and a multiclass SVM for classification to diagnose AMD together with other DR complications. In this method, fundus images first undergo a preprocessing procedure that includes normalization, contrast adjustment or histogram equalization. Then, features are extracted using unsupervised DBN, the dimensions of the feature space are reduced with a generalized regression neural network (GRNN) and finally classification is performed using a multiclass SVM. They used the ARIA dataset to assess the method.
 
 The comparison of these CNN and DBN-based methods given in Table \ref{EM}  shows that CNN-based methods outperform DBN-based methods. Among the CNN-based methods, the one by \citet{abramoff2016improved} achieved the better performance, probably because it is based on an Alexnet-like model. DBNs have not been used in an end-to-end way; thus, they must be explored further using end-to-end learning. Interestingly, DBNs involve significantly fewer learnable parameters than CNN models.

\begin{table}[H]
	\centering
	\caption{Representative works for DME detection}
	\label{EM}
	\hspace*{-4.5cm}	\begin{tabular}{|p{2.3cm}|p{3.3cm}|p{1.5cm}|p{2.5cm}|p{2.2cm}|l|}
		\hline
	Research Study              & Method &Training& Lesion Type(s) & Dataset &Performance\\ \hline
	\multicolumn{6}{|c|}{\cellcolor[HTML]{EFEFEF}CNN-Based Methods} \\ \hline
	%%%%%%%%%%%%%%%%%%%%%%%

		\citet{abramoff2016improved}&CNN inspired by AlexNet&End-to-end& Multistage DR/ME &\begin{tabular}[c]{@{}l@{}}MESSIDOR-2\end{tabular}& SN=100    \\ \hline
		 
		%%%%%%%%%%%%%%%%%%%%%%%%%
			\multirow{2}{*}{\citet{mo2018exudate}}&	\multirow{2}{*}{Cascaded FCRN}&\multirow{2}{*}{End-to-end} &\multirow{2}{*}{ME}&HEI-MED&SN=92.55,F=84.99\\ \cline{5-6}
			
			&&&&e-ophtha&SN=92.27, F=90.53\\ \hline
			%%%%%%%%%%%%%%%%%%%%%%%%%%%%%

		\citet{perdomo2016novel} & Patches based CNN model&Transfer learning&  Multistage DR/ME&MESSIDOR &\begin{tabular}[c]{@{}l@{}}SN=56.5, SP=92.8\\DME ACC=77\\ DME loss=0.78 \end{tabular}
		\\ \hline		
		%%%%%%%%%%%%%%%%%%%

		\citet{burlina2016detection} &CNN-based on OverFeat&Transfer learning   & Multistage AMD& NIH AREDS  & \begin{tabular}[c]{@{}l@{}}SN=90.9-93.4\\SP=89.9-95.6\\ACC=92-95\end{tabular}  \\ \hline
				%%%%%%%%%%%%%%%%%%%%%%%%%%%%%%%%%%%%%%%%%		
	
		\citet{al2016diabetic}&CNN model with three conv. blocks and one FC block&End-to-end& Multistage DR/ME&MESSIDOR&\begin{tabular}[c]{@{}l@{}}SN=74.7, SP=95\\ACC=88.8\end{tabular}\\ \hline
		%%%%%%%%%%%%%%%%%%%%%%%%%%%%%%%%%%%%
		 \citet{ting2017development} &CNN&End-to-end &AMD&35948 images&\begin{tabular}[c]{@{}l@{}}SN=93.2, SP=88.7\\ AUC=0.931\end{tabular}\\ \hline
		 
	%%%%%%%%%%%%%%%%%%%%%%%%%%%%%%

		\multicolumn{6}{|c|}{\cellcolor[HTML]{EFEFEF}DBN-Based Methods} \\ \hline

\citet{arunkumar2017multi}& DBN for training and multiclass SVM as classifier&End-to-end  & AMD& ARIA&\begin{tabular}[c]{@{}l@{}}SN=79.32, SP=97.89\\ACC=96.73\end{tabular}  \\ \hline
		
	\end{tabular} \hspace*{-4.5cm}
\end{table}
%%%%%%%%%%%%%%%%%%%%%%%%%%%%%%%%%%%%%%%%%	

\subsubsection{Exudate as a Clinical Feature }

The detection of EX is necessary for automatic early DR diagnosis, but it is challenging because of significant variation in their size, shape and contrast levels. In this section, we review DL-based methods for EX detection. According to our best knowledge, all of the methods are based on CNNs.

\paragraph{Convolutional Neural Networks}

\citet{prentavsic2016detection} proposed a CNN-based method for EX detection in color fundus images. First, they detect the OD, create an OD probability map and fit a parabola. Then they create vessel probability and bright-border probability maps. Finally, using an 11-layer CNN model, they create an EX probability map and combine it with the OD, vessel and bright-border probability maps and the fitted parabola to generate the final EX probability map. They assessed the model’s performance using the DRiDB database. It significantly outperformed the traditional methods based on hand-engineered features.

\citet{perdomo2017convolutional} proposed a patch-level method based on the LeNet model to discriminate EX regions from healthy regions on fundus images. In this method, potential EX patches are first cropped manually or automatically; then, these patches are passed to the LeNet model for classification. To train LeNet, extra patches are created using a data-augmentation technique based on flipping and rotation operations. The e-ophtha dataset was used for validation; 20,148 EXs and healthy patches were extracted, and 40\% of these patches was used as testing data.

\citet{Gondaletal17} introduced a method for detecting EXs together with other DR lesions based on the award-winning $o\_O$ CNN architecture \cite{oO}. To localize DR lesions, including hard EX (HE) and soft EX (SE), the dense layers are removed from the CNN model. A global average pooling (GAP) layer is introduced on top of the last convolutional layer and is followed by a classification layer, which are used to learn the class-specific importance of each feature map of the last convolution layer. The feature maps are combined with class-specific importance to generate a class activation map (CAM) \cite{zhou2016learning}, which is up-sampled to the size of the original image to localize the lesion regions. The authors used Kaggle for training and DIARETDB1 for validation. This method not only performs image-level detection but also lesion-level detection

\citet{quellec2017deep} addressed the problem of jointly detecting referable DR at the image level and detecting DR lesions such as EXs at the pixel level, and they proposed a solution that relies on CNN visualization methods. The heatmaps generated by CNN visualization techniques are not optimized for computer-aided diagnosis of DR lesions. Based on the sensitivity analysis by \citet{simonyan2014very}, they proposed modifications to generate heatmaps, which help in jointly detecting referable DR and lesions by jointly optimizing CNN predictions and the produced heatmaps. They employed the $o\_O$ architecture as the CNN base model. The authors used the Kaggle dataset for training at the image level and DIARETDB1 for testing at both the lesion and image levels for EX detection.

   \citet{khojasteh2018exudate} compared several DL-patch-based methods to detect EX. They concluded that pre-trained ResNet-50 with SVM outperformed other methods. They assessed their method on DIARETDB1 and e-ophtha.

Table \ref{EX} presents a comparative analysis of the aforementioned methods for EX detection using a DL approach. The methods of  \citet{quellec2017deep,Gondaletal17}, which jointly detect referable DR and lesions, show good performance for both lesion and image-base detection. The method in\citet{khojasteh2018exudate} is computationally more efficient and produces comparable results due to using the deep pre-trained ResNet.

\begin{center}	
	\begin{table}[H]
		\centering
		
		\caption{Representative of works in diabetic retinopathy (DR) EX detection}
		\label{EX}

		\hspace*{-5.5cm}	\begin{tabular}{|p{2.8cm}|p{2.3cm}|l|p{2.1cm}|p{2.1cm}|l|p{2.8cm}|}
			\hline
			Research Study & Method &Training                                                                                                                                                                      & Lesion Type(s)                                                                         & Dataset &\begin{tabular}[c]{@{}l@{}}Segment/\\localize?\end{tabular}&Performance \\ \hline
			
			%%%%%%%%%%%%%%%%%%%%%%%%%%%%%%%%%%%%%%%%%%%%%%%%%%%%
			\multicolumn{7}{|c|}{\cellcolor[HTML]{EFEFEF}CNN-Based Methods} \\ \hline
			%%%%%%%%%%%%%%%%%%%%%%%%%%%%%%%%%%%%%%%%%%%%%%%%%%
				\citet{prentavsic2016detection}&11-layer CNN, OD and vessel maps&End-to-end                                                                                                                                                               & EX&DRiDB &\checkmark& SN=78, F=78
				\\ \hline
				%%%%%%%%%%%%%%%%%%%%%%%%%%%%%%%%%%%%%%%%%%%%%%%%%

		\citet{perdomo2017convolutional}&Patches-based LeNet CNN&Transfer learning&EX&e-ophtha(40\% of patches)&\checkmark &\begin{tabular}[c]{@{}l@{}}SN=99.8, SP=99.6\\ACC=99.6\end{tabular}\\ \hline
	%%%%%%%%%%%%%%%%%%%%%%%%%%%%%%%%%%

				\multirow{2}{*}{\citet{Gondaletal17}} & \multirow{2}{*}{\begin{tabular}[c]{@{}l@{}}$o\_O$ CNN model\\ with CAM\end{tabular}}& \multirow{2}{*}{Transfer learning} &HE/SE
			
				& \multirow{2}{*}{DIARETDB1} &\checkmark& SN:HE=87, SE=80\\ \cline{4-4} \cline{6-7} 
			&                                 &                                    &HE/SE
			&                        &   \xmark &\begin{tabular}[c]{@{}l@{}}SN:HE =100,\\SE=90.0 \\AUC=0.954\end{tabular} \\ \hline
			
					%%%%%%%%%%%%%%%%%%%
				\multirow{2}{*}{\citet{quellec2017deep}} & $o\_O $ CNN(net A)& \multirow{2}{*}{Transfer learning} &HE/SE
				& \multirow{2}{*}{DIARETDB1} &\checkmark &AUC:HE=0.735, SE=0.809\\ \cline{2-2}\cline{4-4} \cline{6-7} 
				&  $o\_O $ CNN(net B)&    &HE/SE
				 &                           & \xmark  & AUC:HE=0.974, SE=0.963 \\ \hline				 
		%%%%%%%%%%%%%%%%%%%%%%
			\multirow{2}{*}{}\citet{khojasteh2018exudate}                    &  	\multirow{2}{*}{\begin{tabular}[c]{@{}l@{}}Patch-based\\ ResNet/SVM as\\ classifier\end{tabular}}&	\multirow{2}{*}{Transfer learning} &EX                                                                                                                                                                   & DIARETDB1 &\xmark &\begin{tabular}[c]{@{}l@{}}SN=99, SP=96, \\ACC=98.2 \end{tabular} \\ \cline{5-7} 
			&    &   &                                                                                                                                                                   & e-ophtha &\xmark  &\begin{tabular}[c]{@{}l@{}}SN=98, SP=95, \\ACC=97.6 \end{tabular}\\ \hline
		
		\end{tabular} \hspace*{-6.5cm}
		
	\end{table}
\end{center}

\subsubsection{Microaneurysms and Hemorrhages as Clinical Features}

MAs and HMs are also have been investigated using DL approaches as a sign of DR, as presented in this section.

 \paragraph{Convolutional Neural Networks (CNN)}
 \citet{haloi2015improved} employed a nine-layer CNN model with a dropout training procedure to classify each pixel as MA or non-MA. Each pixel is classified by taking a window of size 129$\times$129 around it and passing the window to the CNN model. For training, the author employed a data-augmentation technique to generate six windows around each pixel. He graded the severity from no DR to severe DR according to the number of MAs. The method was tested on the MESSIDOR and Retinopathy Online Challenge (ROC) datasets.
 
 The method introduced by \citet{van2016fast} was aimed at detecting HMs. The main contribution of this method is to address the over-represented normal samples created for training a CNN model. To overcome this problem, the authors proposed a dynamic selective sampling strategy that selects informative training samples. First, they extract patches of size 41$\times$41 around HM pixels from positive images only and non-HM pixels from positive images only, and each patch is labeled according to the central pixel. The CNN is trained using a dynamic selective sampling strategy. They used a 10-layer CNN model and tested their system on Kaggle and MESSIDOR.
 
 The methods of \citet{Gondaletal17}  and  \citet{quellec2017deep} discussed in the Exudate section, which jointly detect referable DR and lesions, also detect HMs and small red dots. Another similar method was proposed by \citet{orlando2018ensemble}. In this method, they first extract candidate red lesions using morphological operations and crop patches of size 32×32 around the candidates. Next, they extract CNN features and hand-engineered features (HEFs) such as intensity and shape features from each candidate patch, fuse them and pass the fused feature vector to random forest (RF) to create a probability map, which is used to make lesion- and image-level decisions about red lesions. They employed a six-layer CNN model. For lesion-based evaluation, they used as a competition metric (CPM) the average per lesion sensitivity at the reference false positive detections per image value. They used the DIARETDB1 and e-ophtha datasets for per lesion evaluation. They used MESSIODR for detecting referable DR.
 
 %%%%%%%%%%%%%%

\paragraph{Stacked Autoencoder-Based Methods}
\citet{shan2016deep} used the stacked sparse autoencoder (SSAE) to detect MA lesions. A patch is passed to SSAE, which extracts features, and the Softmax classifier labels it as a MA or non-MA patch. They trained and fine-tuned the SSAE on MA and non-MA patches taken from 89 fundus retinal images selected from the DIARETDB dataset. The patches were extracted without any preprocessing procedure, and Shan and Li evaluated them using 10-fold cross validation.
 
A summary of the above-reviewed methods is given in Table \ref{MA}. In terms of sensitivity, specificity, AUC and accuracy, the CNN-based technique by \citet{haloi2015improved} seems to outperform other methods for MA detection due to using pixel augmentation instead of image-based augmentation. The performance of the stacked sparse autoencoder based-method by \citet{shan2016deep} is not better than those based on CNN. Among CNN-based methods, those used by \citet{Gondaletal17}  and \citet{quellec2017deep} are computationally efficient and jointly detect referable DR and red lesions. 
	
 \begin{table}[]

 	\caption{Representative of works in diabetic retinopathy (DR) detection based on MA and HM}
 	\label{MA}

\centering 	 	 	
 \begin{tabular}{|p{2.7cm}|p{2.0cm}|p{1.6cm}|l|p{2.3cm}|l|p{2.9cm}|}

 	\hline
 	\begin{tabular}[c]{@{}l@{}}Research\\study\end{tabular} & Method&Training&Lesion Type(s)&Dataset &   \begin{tabular}[c]{@{}l@{}}Segment/\\localize?\end{tabular}&Performance\\ \hline 	
 	
 	\multicolumn{7}{|c|}{\cellcolor[HTML]{EFEFEF}CNN Based Methods} \\ \hline
 	%%%%%%%%%%%%%%%%%%%%%%%%%%%%%%%%%%%%%%
 	
 	\multirow{2}{*}{\citet{haloi2015improved}}&\multirow{2}{*}{9-layer CNN}&\multirow{2}{*}{End-to-end}& \multirow{2}{*}{MA}& MESSIDOR &\checkmark&\begin{tabular}[c]{@{}l@{}}SN=97, SP=95\\AUC=0.982\\ACC=95.4\end{tabular}\\ \cline{5-7} 
 	&                                                                                                  &                                                                               &                                                                               &ROC  &\checkmark&AUC=0.98\\ \hline
 	%%%%%%%%%%%%%%%%%%%%%%%%%%%%%%%%%%
 	\multirow{2}{*}{\begin{tabular}[c]{@{}p{2.3cm}@{}}\citet{van2016fast}\end{tabular}}            & \multirow{2}{*}{\begin{tabular}[c]{@{}l@{}}Patches based \\ selective\\ sampling\end{tabular}}     & \multirow{2}{*}{End-to-end}                                                   & \multirow{2}{*}{HM}                                                           & Kaggle &\checkmark&\begin{tabular}[c]{@{}l@{}}SN=84.8, SP=90.4\\AUC=0.917\end{tabular}\\ \cline{5-7} 
 	&                                                                                                  &                                                                               &                                                    & MESSIDOR&\checkmark&\begin{tabular}[c]{@{}l@{}}SN=93.1, SP=91.5\\AUC=0.979\end{tabular}\\ \hline
 	%%%%%%%%%%%%%%%%%%%%%%%%%%%%%%%%%%%%%%%%%
 	
 	\multirow{2}{*}{\citet{Gondaletal17}}                  & \multirow{2}{*}{\begin{tabular}[c]{@{}l@{}}o\_O CNN\\ model\end{tabular} }                                                                  & \multirow{2}{*}{\begin{tabular}[c]{@{}l@{}}Transfer\\ learning\end{tabular}}  & \begin{tabular}[c]{@{}l@{}}-HM\\-Small red dots\end{tabular} & \multirow{2}{*}{DIARETDB1} &\checkmark& \begin{tabular}[c]{@{}l@{}}SN:\\-HM=91\\ -Small red dots=52\end{tabular}\\ \cline{4-4} \cline{6-7} 
 	&                                                                                                  &                                                                               & \begin{tabular}[c]{@{}l@{}}-HM\\-Small red dots\\ \end{tabular} &                                                                                          &\xmark  & \begin{tabular}[c]{@{}l@{}}SN:\\-HM=97.2\\ -Red small dots=50\end{tabular}\\ \hline
 	%%%%%%%%%%%%%%%%%%%%%%%%%%%%%%%%%%%%%%%%%
 	\multirow{2}{*}{\citet{quellec2017deep}}                  & \multirow{2}{*}{\begin{tabular}[c]{@{}l@{}}o\_O CNN\\(net B)\end{tabular}}                                                                 & \multirow{2}{*}{\begin{tabular}[c]{@{}l@{}}Transfer\\  learning\end{tabular}} & \begin{tabular}[c]{@{}l@{}}-HM\\ -Small red dots\\ \end{tabular} & \multirow{2}{*}{DIARETDB1}   &\checkmark                                                              & \begin{tabular}[c]{@{}l@{}}AUC:\\-HM=0.614\\-Small red dots=0.50\end{tabular}\\ \cline{4-4} \cline{6-7} 
 	&                                                                                                  &                                                                               & \begin{tabular}[c]{@{}l@{}}-HM\\-Small red dots\end{tabular} &&\xmark& \begin{tabular}[c]{@{}l@{}}AUC:\\-HM=0.999\\-Small red\\dots=0.912\\-Red small dots\\+HM=0.97\end{tabular}                   \\ \hline
 	\multirow{4}{*}{\citet{orlando2018ensemble}}&\multirow{3}{*}{\begin{tabular}[c]{@{}l@{}}HEF + \\CNN features\\  and RF\\classifier\end{tabular}} & \multirow{3}{*}{End-to-end}                                                   & MA& \multirow{2}{*}{DIARETDB1}  &\checkmark &CPM=0.3301, SN=48.83\\ \cline{4-4} \cline{6-7} 
 	&                                                                                                  &                                                                               & HM                                                         &                                                                                          & \checkmark&CPM=0.4884, SN=48.83\\ \cline{4-7} 
 	
 		&                                                                                                  &                                                                               &\begin{tabular}[c]{@{}l@{}}MA\end{tabular}                      & e-ophtha  &\checkmark                                                            &\begin{tabular}[c]{@{}l@{}}CPM=0.3683,\\SN=36.80\end{tabular}\\ \cline{4-7} 
 	
 	&                                                                                                  &                                                                               &\begin{tabular}[c]{@{}l@{}}Red lesion\end{tabular}                      & MESSIDOR  &\xmark                                                            &\begin{tabular}[c]{@{}l@{}}SN=91.09, SP=50\\AUC=0.8932\end{tabular}\\ \hline
 	
 	\multicolumn{7}{|c|}{\cellcolor[HTML]{EFEFEF}AE Based Methods} \\ \hline
 	%%%%%%%%%%%%%%%%%%%%%%%%%%%%%%%%%%%%
 	
 	\citet{shan2016deep} & \begin{tabular}[c]{@{}l@{}}Patches based\\ SSAE\end{tabular}& \begin{tabular}[c]{@{}l@{}}Transfer\\ learning\end{tabular}& MA&DIARETDB&\checkmark&\begin{tabular}[c]{@{}l@{}}SP=91.6\\F=91.3\\ACC=91.38\end{tabular}\\ \hline
 \end{tabular}			

 \end{table}

 \subsection{Classification of Fundus Images for Referral}
 This section focuses on methods that deal with referable DR detection and use only image-level annotation. The main purpose of these methods is to grade DR levels for referral. Some methods in this category also detect lesions jointly with referable DR detection, but without using pixel- or lesion-level annotation \cite{quellec2017deep}. To the best of our knowledge, only CNN models have been employed for this problem.
 
\citet{gulshan2016development} used the Inception-v3 CNN architecture to detect referable DR on a fundus image. They assessed the system using the EyePACS-1 dataset, which consists of 9,963 images taken from 4,997 patients and MESSIDOR-2; both were graded by at least seven US licensed ophthalmologists and ophthalmology senior residents. This evaluation study concluded that an algorithm based on CNN has high sensitivity and specificity for detecting referable DR. 

\citet{colas2016deep} proposed a method based on deep learning, which jointly detects referable DR and lesion location. They trained the deep model on 70,000 labeled images and tested 10,000 images taken from Kaggle dataset, where each patient has two images of right and left eyes. Each image is graded by ophthalmologists into five main stages that vary from no retinopathy to proliferative retinopathy.  

Similarly, as discussed in the MAs and HMs section, the method used by \citet{quellec2017deep} jointly detects referable DR and lesions; its performance was evaluated on three datasets: Kaggle, e-ophtha and DIRETDB1.

\citet{costa2017convolutional} used a different approach and introduced a method for detecting referable DR by generalizing the idea of bag-of-visual-words (BoVW). First, they extract sparse local features with speeded-up robust features (SURF) and encode them using convolution operation or encoded dense features with a CNN model, and then use neural network for classification. They evaluated the proposed methods on three different datasets: DR1 and DR2 from \cite{pires2014advancing} and MESSIDOR. DR1 and DR2 consist of grayscale images. The authors show that the SURF-based method outperforms the CNN-based method, probably because the CNN architecture is not deep enough.

\citet{pratt2016convolutional} used CNN structure for grading fundus images into one of the five stages: no DR, mild DR, moderate DR, severe DR and proliferated DR. They addressed the issues of overfitting and skewed datasets, and proposed a technique to solve these issues. For training, they enhanced the volume of data using a data augmentation technique. The employed CNN model consists of 10 convolutional layers and three fully connected layers. For training, they used 80,000 images taken from the Kaggle dataset, and 5,000 images for testing.  

\citet{gargeya2017automated} used the ResNet CNN model consisting of five residual blocks of four, six, eight, 10, and six layers, respectively, and a gradient boosting classifier for grading a fundus image as normal or referable DR. Additionally, they introduced a convolutional visualization layer at the end of ResNet for visualizing its learning procedure. For training, they used 75,137 images selected from the EyePACS dataset and evaluated independently on MESSIDOR-2 and e-ophtha datasets.

\citet{abramoff2016improved} also proposed a method based on supervised end-to-end CNN models, discussed in the ME section, to grade a fundus image as normal or referable DR; a fundus image is taken to be referable DR if it is moderate DR, severe non-proliferative DR (NPDR) or proliferative DR (PDR).

Similarly, \citet{ting2017development} addressed the problem of detecting referable DR using 76,370 images for training. This method is discussed in the mecula edema section.

\citet{wang2017zoom} proposed a supervised image-level CNN-based approach that diagnosed DR and highlighted suspicious patches regions. They used a network called Zoom-in, which mimics the zoomin procedure of retinal clinical examination. The architecture of the network consisted of three parts: main network (M-Net), which was pre-trained on ImageNet, a sub-network; attention network (A-Net) to generate attention maps; and another sub-network, crop-network (C-Net). They used EyePACS and MESSIDOR to evaluate the system.

The method by \citet{mansour2018deep} used the AlexNet model in conjunction with a preprocessing, Gaussian mixture model for background subtraction and connected component analysis to localize blood vessels; then linear discriminant analysis is used for dimensionality reduction. Finally, SVM is employed for classification and 10-fold cross validation is used for evaluation. Also, as discussed in the MAs and HMs section, the method by \citet{orlando2018ensemble}, jointly detects referable DR and lesions; its performance was evaluated on MESSIDOR.

\citet{chen2018diabetic} built a model called SI2DRNet-v1 to detect referral DR that consisted of 20 layers. After applying preprocessing in their model, such as a Gaussian filter, they used global average pooling instead of FC layers and 1$\times$1 filters to reduce parameters and regularize the model; they also scaled the kernel size after each pooling layer from 3$\times$3 to 5$\times$5. Finally, they extracted 5 probability values from the Softmax layer to grade DR severity.
 
 Table \ref{HL} summarizes the works presented in this section. Apparently, the method by \citet{gulshan2016development} outperforms other methods in terms of sensitivity, specificity and AUC; the performance of this method is comparable to a panel of seven certified ophthalmologists. However, it is difficult to compare the performance of the methods because different datasets were used for training and testing. The method by \citet{gargeya2017automated} seems to be robust because it is based on ResNet architecture, which has been shown to outperform most of the CNN architectures; it was trained and tested on different datasets, and its cross-dataset performance is quite good. A CNN model is like a black box and does not give any insight into pathology. This method also incorporates the visualization of pathologic regions, which can aid real-time clinical validation of automated diagnoses.
 %%%%%%%%%%%%%%%%%%%%%%%%%%%
	   \begin{table} [H]
	    	\caption{Representative of works in diabetic retinopathy (DR) for referral }
	    	\label{HL}
	    \centering
	    \hspace*{-5cm}	%\begin{tabular}{|p{2cm}|p{2.5cm}|p{2.5cm}|p{3cm}|p{1.5cm}|p{1.5cm}|p{1cm}|p{1cm}|l|l|}
	    \begin{tabular}{|l|p{2.5cm}|l|l|p{1cm}|p{1cm}|p{0.9cm}|p{1cm}|}
	    		%\centering

	    		\hline
	    		Research Study& Method&Training&Dataset& SN\%& SP\%&AUC                                                                   & ACC\% \\ \hline
%%%%%%%%%%%%%%%%%%%%%%%%%%%%%%%

	    			\multirow{2}{*}{\citet{gulshan2016development} }           & 	\multirow{2}{*}{Inception-v3 CNN}& 	\multirow{2}{*}{Transfer learning}                                                     &  EyePACS-1 & 90.3   &90& 0.991                                                                 & -     \\ \cline{4-8} 
	    			
	    			&             &   & MESSIDOR-2& 87      & 98.5       & 0.990                                                                 & -     \\ \hline
	    			%%%%%%%%%%%%%%%%%%%%%%%%%%%%%%
	    				\citet{colas2016deep}                     & CNN model&End-to-end& Kaggle                                                                                & 96.2        & 66.6       &   0.946& -     \\ \hline
	    				
	    				%%%%%%%%%%%%%%%%%%%%%%%%

	    		\multirow{3}{*}{\citet{quellec2017deep}}
	    		& $o\_O $ CNN(net B)& 	\multirow{3}{*}{Transfer learning}                                                         & DIARETDB1                                                                               & -           & -           & 0.954                                                                 & -     \\ \cline{2-2} \cline{4-8}

	    		  & 	\multirow{2}{*}{\begin{tabular}[c]{@{}p{2.5cm}@{}}Ensemble net A, net B, AlexNet\end{tabular}} && Kaggle                                                                                 & -           & -           & 0.955                                                                 & -     \\ \cline{4-8} 
	    		&                                                          &   & e-ophtha                                                                              & -           & -           & 0.949                                                                 & -     \\ \hline
	    		%%%%%%%%%%%%%%%%%%%%%%%%%%%%%%%%%%
	    	
	    			\multirow{3}{*}{\citet{costa2017convolutional}}
	    			& 	\multirow{3}{*}{\begin{tabular}[c]{@{}l@{}} Sparse\\ SURF/CNN\end{tabular}}          &	\multirow{3}{*}{ End-to-end}	&  MESSIDOR(20\% of images )                                                                        & -        &-       & 0.90 & -  \\ \cline{4-8} &
	    		         &   &	   DR1(20\% of images )                                                                          & -        &-       & 0.93 & -  \\ \cline{4-8}
	    			
	    			& &    &   DR2(20\% of images)                                                                          & -        &-       & 0.97 & -  \\ \hline
	    		
	    		%%%%%%%%%%%%%%%%%%%%%%%%%%%%%%
	    	
	\citet{pratt2016convolutional}&13-layers CNN& End-to-end & Kaggle& 95&-& -& 75\\ \hline		    
	
	%%%%%%%%%%%%%%%%

\multirow{2}{*}{\citet{gargeya2017automated}} &\multirow{2}{*}{\begin{tabular}[c]{@{}p{2.8cm}}ResNet+Gradient boosting  tree\end{tabular}}&\multirow{2}{*}{End-to-end} &  MESSIDOR-2& - & - & 0.94 & - \\ \cline{4-8} 
&  &  &  \begin{tabular}[c]{@{}l@{}} e-ophtha\end{tabular}&- & - & 0.95 &- \\ \hline

	%%%%%%%%%%%
	\citet{ting2017development} & CNN&End-to-end&(71896 images) &90.5&91.6&0.936&-\\ \hline
	%%%%%%%%
	\citet{abramoff2016improved}&CNN& End-to-end&\begin{tabular}[c]{@{}l@{}}MESSIDOR-2\end{tabular}& 96.8& 87&0.980& -     \\ \hline
	%%%%%%%%%%%%%%%%%%%%%%%%%%%%%%%%%%%%
\multirow{2}{*}{\citet{wang2017zoom} }           & 	\multirow{2}{*}{Zoom-in network}& 	\multirow{2}{*}{Transfer learning}                                                     &  EyePACS & -  &-& 0.825                                                                 & -     \\ \cline{4-8} 

&             &   & MESSIDOR& -      & -       & 0.957 &91.1                                                                    \\ \hline
	%%%%%%%%%%%%%%%%%%%%%%%%%
	\citet{chen2018diabetic}&SI2DRNet-v1(20 layers)&End-to-end&MESSIDOR&-&-& 0.965&91.2\\ \hline
	%%%%%%%%%%%%%%%%%%%%%%%%
	\citet{mansour2018deep}&AlexNet/SVM as classifier&Transfer learning&Kaggle&100&93&-&97.93\\ \hline
	%%%%%%%%%%%%%%%%%%%%%%%%%%%%%%
	\citet{orlando2018ensemble}&\begin{tabular}[c]{@{}l@{}}HEF + \\CNN features\\  and RF\\classifier\end{tabular} & End-to-end&MESSIDOR& 97.21 & 50         & 0.9347 &-            \\ \hline                                                

	\end{tabular} \hspace*{-5cm}
	 \end{table}

\section{Discussion}\label{DIS}

The previous section gives a detailed account of techniques related to DR diagnosis based on a deep-learning approach. The studies listed in this survey use four main deep learning architectures: CNN, AE, DBN and RNN. Each of these architectures has several variations that have been used in DR diagnosis. Deep-learning-based techniques have been proposed for retinal vessels segmentation, OD detection and segmentation, DR lesion detection and classification, and referable DR detection. A review of these methods indicates that most deep-learning-based techniques for the above problems use CNN architecture, and that it outperforms other deep architectures. Despite these improvements, there are still challenges to improve deep learning techniques for more robust and accurate detection, localization and diagnosis of different DR biomarkers and complications. All reviewed methods were tested and evaluated on public domain datasets, except two methods \cite{srivastava2015using,ting2017development} that used fundus images collected from medical organizations and hospitals. For the most part, methods addressing the same problem were evaluated on different datasets using different metrics, and as such, it is difficult to precisely compare them and grade them based on their performance. Most of the methods were evaluated using the same dataset for training and testing, and performance of the same method is different for different datasets; this raises questions about their robustness and how these methods will perform when deployed in a real clinical setting.

There is a serious difficulty in interpreting and comparing the results of different methods in terms different performance metrics when different datasets are used for evaluation. For example, method by \citet{liskowski2016segmenting} is tested on STARE consisting of 402 images (38 test negatives and 364 test positives) and method by \citet{dasgupta2017fully} is tested on the DRIVE consisting of 40 images (33 test negatives and 7 test negatives). Both methods are similar in terms of specificity (SP),  area under ROC curve (AUC) and accuracy (ACC) i.e. both have  SP = 98\%,  AUC = 0.97-0.99 and ACC = 95\%-97\%., but the method by \citet{liskowski2016segmenting} is far better (with SN = 85\%) than the method by \citet{dasgupta2017fully} (with SN = 76\%) when sensitivity is used for evaluation. It indicates that the methods must be evaluated on the same datasets to estimate their real performance gains. 

Though CNN architecture results in better performance, CNN models involve a huge number of parameters, and requires a huge volume of annotated datasets; however, the available datasets consist of a small number of annotated images. As such, when CNN is used to detect and diagnose different DR complications, there is a high risk of overfitting. One solution to deal this problem is data augmentation, but the data augmentation techniques that have been used so far do not create real samples. More data augmentation techniques are needed in order to create new samples from existing ones. Another solution is to use transfer learning, i.e. first train a CNN model using a dataset from a related domain and then fine-tune it with the dataset from the domain of the problem. Most of the reviewed methods use CNN models pre-trained on natural images \cite{maninis2016deep,niu2017automatic,xu2017optic,burlina2016detection}, e.g. the ImageNet dataset for transfer learning; only a few methods used CNN models pre-trained on fundus images \cite{quellec2017deep,Gondaletal17}. Another alternative to deal with the overfitting problem is to introduce CNN models that are expressive but involve fewer learnable parameters. 

%%%%%%%%% 
\subsection{Comparison of Deep-Learning-Based and Hand-Engineered Methods}

In this section, we compare traditional methods based on hand-engineered features and deep-learning-based methods. For comparison, we selected the methods evaluated using the same datasets and performance metrics. The selected traditional methods are the state-of-the-art methods reported in references \cite{srinidhi2017recent,almotiri2018retinal} for vessels segmentation and \cite{almazroa2015optic} for OD and \cite{mookiah2013computer} for MAs. The deep learning methods that give the best performance in this review are selected for comparison. Although hand-engineered features have been dominant for long time, the deep learning approach is a state-of-the-art technique and has shown impressive performance compared with traditional approaches. Table \ref{comp} presents a comparison between traditional and deep-learning-based methods.

\begin{figure}[H]
	\centering
	\includegraphics[width=0.7\linewidth]{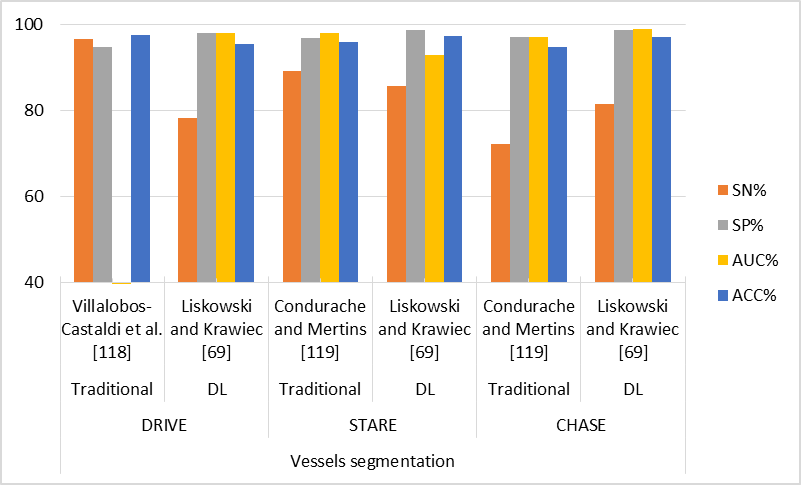}

	\caption{The plot for maximum performance of vessel segmentation in traditional and DL-based methods}
	\label{fig:screenshot041}
\end{figure}

For retinal blood vessel segmentation, the traditional method by \citet{villalobos2010fast} seems to give better sensitivity (96.48\%) and accuracy (97.59\%) than the deep learning method used by \citet{liskowski2016segmenting} (sensitivity: 78.11\% and accuracy: 95.35\%) on the DRIVE dataset; however, the latter method gives overall better performance than another traditional method by \citet{condurache2012segmentation} on the STARE and CHASE datasets. This indicates that deep learning based methods outperform traditional methods overall, but in spite of this fact, even deep-learning-based methods are not robust: their performance is different for different datasets. Figure \ref{fig:screenshot041} shows a plot of the performance of traditional and DL-based methods.

For OD localization, the deep-learning-based method by \citet{zhang2018automatic}, with an accuracy of 99.9\%, outperforms the traditional method used by \citet{aquino2010detecting}, with an accuracy of 99\% on the MESSIDOR dataset. For the DRIVE dataset, the learning-based method by \citet{alghamdi2016automatic} and traditional method by \citet{zhang2012optic} achieved the same performance with an accuracy of 100\%. However, for DIARETDB1, the traditional method by \citet{sinha2012optic}, with an accuracy of 100\%, outperforms the deep-learning accuracy of 98.88\% achieved by \citet{alghamdi2016automatic}.

\begin{figure}[H]
	\centering
	\includegraphics[width=0.7\linewidth]{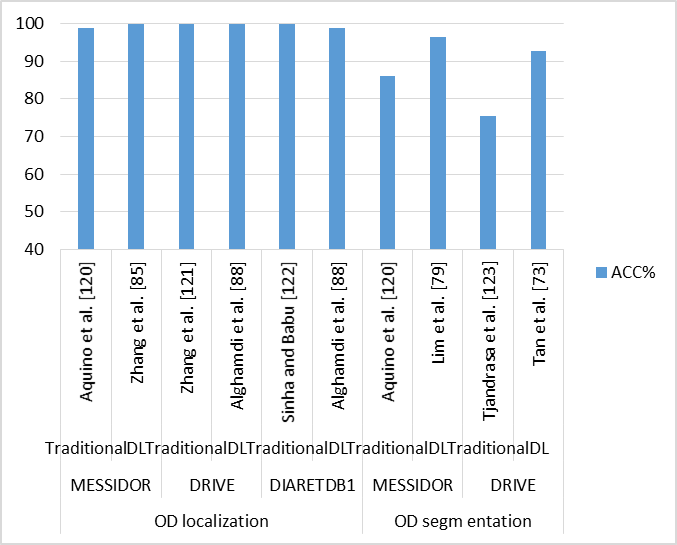}
	\caption{The plot for maximum performance of OD localization and segmentation in traditional and DL-based methods}
	\label{fig:screenshot042}
\end{figure}

For OD segmentation, deep-learning-based methods show significantly higher accuracy than traditional methods. For example, on the MESSIDOR dataset, the traditional method used by \citet{aquino2010detecting} showed less accuracy (86\%) than that of the deep-learning-based method of \citet{lim2015integrated}, achieving significant higher accuracy (96.4\%). Similarly, on the DRIVE dataset, the traditional method used by \citet{tjandrasa2012optic} gave lower accuracy (75.56\%) than that yielded by Tan et al.’s deep-learning-based method. (92.68\%) \citet{tan2017segmentation}. Figure \ref{fig:screenshot042} summarizes the performance of OD localization and segmentation in traditional and DL methods.

Exudate detection review by \citet{joshi2018review} that published 2018 reported the maximum EX detection performance on private datasets. However, we compared the best DL method with the one that has maximum number of images and higher performance. \citet{massey2009robust} showed slightly better accuracy (98.87\%) whereas the DL based method by \citet{khojasteh2018exudate} achieved 98.2\% but with higher sensitivity (99\%).

\begin{figure}[H]
\centering
\includegraphics[width=0.55\linewidth]{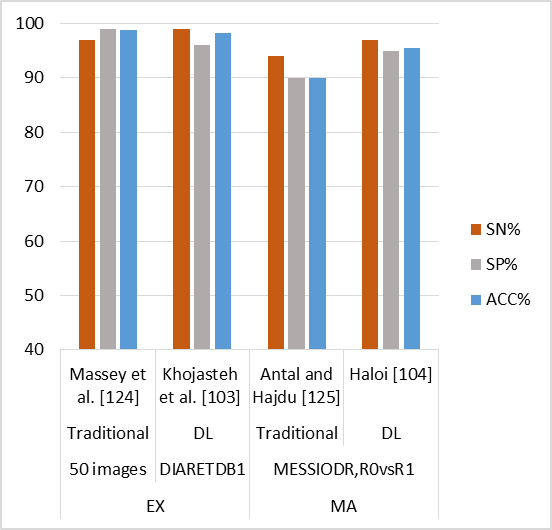}
\caption{The plot for maximum performance of MA and EX detection in traditional and DL-based methods}
\label{fig:screenshot043}
\end{figure}

For MAs detection, on the MESSIDOR dataset, Haloi’s deep-learning-based method \cite{haloi2015improved} achieved a higher performance with sensitivity, specificity, AUC and accuracy of 97\%, 95\%, 0.982 and 95.4\%, respectively, whereas the traditional method by \citet{antal2012ensemble} equaled 94\%, 90\%, 0.942 and 90\%, respectively.  Figure \ref{fig:screenshot043} presents both EX and MA detection the performance in traditional and DL methods.

Overall, CNN-based methods for retinal vessel segmentation, OD detection and segmentation and DR lesion detection outperform traditional methods. However, deep-learning methods are not robust, not interpretable and suffer from overfitting, and further research is needed to overcome these issues. 

%%%%%%%%%%%%%%%%%%%%%
\begin{table}[H]
	\caption{Comparison between state-of-the-art traditional methods and best-performance deep-learning methods}
	\label{comp}
	\begin{tabular}{|l|l|l|l|l|l|l|l|}
		\hline
		Features                                                                    & Approach     & Study                                           & Dataset                                                              & SN\%                                                              & SP\%    & AUC                                                                      & ACC\%  \\ \hline
		\multirow{6}{*}{Vessels}                                                    & Traditional &  \citet{villalobos2010fast}        & \multirow{2}{*}{DRIVE}                                               & 96.48                                                           & 94.80 & -                                                                   & 97.59  \\ \cline{2-3} \cline{5-8} 
		&\cellcolor[HTML]{FFEFFF} Deep learning&\cellcolor[HTML]{FFEFFF}\citet{liskowski2016segmenting} &                                                                      & \cellcolor[HTML]{FFEFFF}78.11                                                            &\cellcolor[HTML]{FFEFFF} 98.07 & \cellcolor[HTML]{FFEFFF}0.9790                                                                   &\cellcolor[HTML]{FFEFFF} 95.35  \\ \cline{2-8} 
		& Traditional &  \citet{condurache2012segmentation}        & \multirow{2}{*}{STARE}                                               & 89.02& 96.73& 0.9791 & 95.95 \\ \cline{2-3} \cline{5-8} 
		&\cellcolor[HTML]{FFEFFF} Deep learning          &  \cellcolor[HTML]{FFEFFF}\citet{liskowski2016segmenting} &                                                                      & \cellcolor[HTML]{FFEFFF}85.54                                                            & \cellcolor[HTML]{FFEFFF}98.62 & \cellcolor[HTML]{FFEFFF}0.9928                                                                   & \cellcolor[HTML]{FFEFFF}97.29  \\ \cline{2-8} 
		& Traditional &  \citet{condurache2012segmentation}        & \multirow{2}{*}{CHASE}                                               & 72.24                                                            & 97.11 &0.9712                                                                    &  94.69 \\ \cline{2-3} \cline{5-8} 
		&\cellcolor[HTML]{FFEFFF} Deep learning          &  \cellcolor[HTML]{FFEFFF}\citet{liskowski2016segmenting} &                                                                      & \cellcolor[HTML]{FFEFFF}81.54                                                            & \cellcolor[HTML]{FFEFFF}98.66 & \cellcolor[HTML]{FFEFFF}0.988                                                                    &\cellcolor[HTML]{FFEFFF} 96.96  \\ \hline

				\multirow{6}{*}{\begin{tabular}[c]{@{}l@{}}OD \\ localization\end{tabular}} & Traditional & \citet{aquino2010detecting} & 	\multirow{2}{*}{MESSIDOR}                                                              & - & - & - & 99 \\ \cline{2-3} \cline{5-8} 
				&\cellcolor[HTML]{FFEFFF} Deep learning          & \cellcolor[HTML]{FFEFFF}\citet{zhang2018automatic}     &    & \cellcolor[HTML]{FFEFFF}- &\cellcolor[HTML]{FFEFFF} - & \cellcolor[HTML]{FFEFFF}- & \cellcolor[HTML]{FFEFFF}99.9 
				\\ \cline{2-8} 
			 & Traditional & \citet{zhang2012optic} & 	\multirow{2}{*}{DRIVE}                                                              & - & - & - & 100 \\ \cline{2-3} \cline{5-8} 
			 & \cellcolor[HTML]{FFEFFF}Deep learning          & \cellcolor[HTML]{FFEFFF}\citet{alghamdi2016automatic}  &    &\cellcolor[HTML]{FFEFFF} - &\cellcolor[HTML]{FFEFFF} - &\cellcolor[HTML]{FFEFFF} - &\cellcolor[HTML]{FFEFFF} 100   
		\\ \cline{2-8} 
		%%%%
		 & Traditional & \citet{sinha2012optic} & 	\multirow{2}{*}{DIARETDB1}                                                              & - & - & - & 100 \\ \cline{2-3} \cline{5-8} 
		 &\cellcolor[HTML]{FFEFFF} Deep learning          & \cellcolor[HTML]{FFEFFF}\citet{alghamdi2016automatic}  &   &\cellcolor[HTML]{FFEFFF} - & \cellcolor[HTML]{FFEFFF}- & \cellcolor[HTML]{FFEFFF}- & \cellcolor[HTML]{FFEFFF}98.88 \\ \hline
		 
		 %%%%%%%%%%%%%%%%%%%%%%%%%%%%%%%%%
		
		\multirow{4}{*}{\begin{tabular}[c]{@{}l@{}}OD segm\\ entation\end{tabular}} & Traditional &  \citet{aquino2010detecting}     & \multirow{2}{*}{MESSIDOR}                                            & -                                                                 & -      & -                                                                        & 86     \\ \cline{2-3} \cline{5-8} 
		&\cellcolor[HTML]{FFEFFF} Deep learning          &  \cellcolor[HTML]{FFEFFF}\citet{lim2015integrated}       &                                                                      & \cellcolor[HTML]{FFEFFF}-                                                                 &\cellcolor[HTML]{FFEFFF} -      & \cellcolor[HTML]{FFEFFF}-                                                                        &\cellcolor[HTML]{FFEFFF} 96.4   \\ \cline{2-8} 
		& Traditional &  \citet{tjandrasa2012optic}      & \multirow{2}{*}{DRIVE}                                               & -                                                                 & -      & -                                                                        & 75.56  \\ \cline{2-3} \cline{5-8} 
		&\cellcolor[HTML]{FFEFFF} Deep learning          &  \cellcolor[HTML]{FFEFFF}\citet{tan2017segmentation}     &                                                                      & \cellcolor[HTML]{FFEFFF}87.90                                                            &\cellcolor[HTML]{FFEFFF} 99.27 & \cellcolor[HTML]{FFEFFF}-                                                                        & \cellcolor[HTML]{FFEFFF}92.68  \\ \hline

		\multirow{2}{*}{EX} & Traditional   & \citet{massey2009robust}   & 50 images%,\\per lesion
				& 96.9& 98.9&-& 98.87 \\ \cline{2-3} \cline{5-8} 
		&\cellcolor[HTML]{FFEFFF} Deep learning & \cellcolor[HTML]{FFEFFF}\citet{khojasteh2018exudate} & DIARETDB1                          &\cellcolor[HTML]{FFEFFF}99&\cellcolor[HTML]{FFEFFF} 96 &\cellcolor[HTML]{FFEFFF}-& \cellcolor[HTML]{FFEFFF}98.2\ \\ \hline
	%%%%%%%%%%%%%%%%%%%%%%%%%%%%%%%%%%%%%%%%%
		\multirow{2}{*}{MA} & Traditional   & \citet{antal2012ensemble}   & \multirow{2}{*}{\begin{tabular}[c]{@{}l@{}}MESSIODR
				 ,R0vsR1\end{tabular}} & 94 & 90 & 0.942   & 90 \\ \cline{2-3} \cline{5-8} 
		&\cellcolor[HTML]{FFEFFF} Deep learning & \cellcolor[HTML]{FFEFFF}\citet{haloi2015improved} &                           &\cellcolor[HTML]{FFEFFF}97&\cellcolor[HTML]{FFEFFF} 95 &\cellcolor[HTML]{FFEFFF} 0.982& \cellcolor[HTML]{FFEFFF}95.4\ \\ \hline
		
	\end{tabular}
\end{table}

\section{Gaps and Future Directions}\label{GAP}
This review of methods related to DR diagnosis reveals that deep learning helped to design better methods for DR diagnosis and moved state-of-the-art techniques forward, but it is still an open problem, and more research is needed. There are not many methods based on deep learning, and advanced deep-learning techniques must be developed in order to solve this problem. Deep-learning-based models are mostly black boxes and do not provide interpretations of diagnostic value that could help validate their usefulness in a real clinical setting. Most of the methods in this review do not provide any interpretation of their outcomes.

One of the most challenging problems in designing robust deep-learning methods, especially based on CNN models with deeper architectures, is the acquisition of huge volumes of labeled fundus images with pixel- and image-level annotations. The main issue is not the availability of huge datasets, but the annotation of these images, which is expensive and requires the services of expert ophthalmologists. The solution is to design learning algorithms that can learn a deep model from limited data; this is an important area of research not only for DR diagnosis, but also in medical image analysis; one possible direction to explore Generative Adversarial Networks (GANs) \cite{neff2017generative}. Another alternative is to introduce augmentation techniques, as the data augmentation techniques used thus far do not create real samples. Therefore, new data augmentation techniques must be developed that create new samples from existing samples that represent real samples. Another alternative to deal with the problem is to introduce CNN models that are expressive but involve fewer learnable parameters.

Moreover, class imbalance is another challenge in datasets; in medical imaging, in general, and fundus images, in particular, the number of DR cases is much less than in normal cases. Furthermore, the quantities of images with different DR complications and DR lesions are different, and this difference in some cases is significant and adds bias for specific classes during the training of deep models. Large-scale retinal screening processes around the world lead to huge datasets of fundus images; however, most of the images are normal and do not contain any suspicious symptoms or lesions. Developing deep-learning strategies in dealing with this class imbalance is another essential area of research. Data augmentation has been used in some studies – such as \cite{pratt2016convolutional,van2016fast,worrall2016automated,chen2015glaucoma} – to tackle the class imbalance problem, but these data augmentation techniques mostly use geometric transformations and only create rotated and scaled samples, and do not introduce samples with lesions having morphological variations. More sophisticated data augmentation techniques that create heterogeneous samples while preserving prognostic characteristics of fundus images must be introduced, and one possible direction is to explore Generative Adversarial Networks (GANs) \cite{frid2018gan}.

A principal issue in fundoscopy is the lack of uniformity among fundus images, i.e. the images being captured under different conditions. Fundus images usually suffer from the problem of illumination variation due to non-uniform diffusion of light in the retina; the shape of a retina is close to a sphere, which prevents the retina’s light incident from being reflected uniformly. Another common problem with respect to illumination is related to the angle at which light is incident on the retina; the angle at which the image is taken is not always the same. This can be confirmed by observing that the optic nerve does not maintain a specific position in the entire database. Another problem related to capturing the fundus image is that in some cases, the image is out of focus. In addition, fundus images are not always captured with the same resolution and camera. There is also the problem of pigmentation reflected by the iris. 

One way to deal with these problems is to add a preprocessing stage in deep-learning methods. Alternatively, a robust approach is to design deep models such as Generative Adversarial Networks (GANs) \cite{schawinski2017generative} so that they automatically detect and correct these image artifacts. The bottleneck herein is to develop a huge annotated dataset that captures all different types of image artifacts. Alternatively, sophisticated data augmentation techniques must be introduced, augmenting an existing dataset with images having different artifacts. 

The methods developed so far are based on the assumption that the input image is a retinal image. However, the input image might not be a true retinal image or tempered retinal image; with the development of user-friendly image editing software it is easy to tamper an image.  As such first of all an intelligent computer aided method must first of all identify whether the input image is a real and authentic retinal image.

Ophthalmologists usually prefer to use pupil dilating drops (mydriasis) for better view of the retina and more field to evaluate if there is peripheral DR. Using fundus photography can be with mydriasis or without depending on the field of camera lens used. Optos camera can capture up to 200$^{\circ}$ of the retina in one shot and without mydriasis, while other cameras like Topcon and Nikon (which is used usually to screen DR as it has better image quality and color) can capture 30-50$^{\circ}$ which is a limitation of the system if we capture one image. \citet{lawrence2004accuracy} conducted a study to compare mydriasis and non-mydriasis with single or multiple shots to clarify how sensitive is mydriasis in DR screening. He showed that mydriasis with multiple shots is more sensitive and specific compared to in non-mydriasis group with single shot. For that reason, the datasets must be developed with mydriasis and multiple shots to avoid the ungradable image. The rate of ungradable images in the general ranges from 7-17\% \cite{massin2003evaluation,szabo2015telemedical,abdellaoui2016screening,siu1998effectiveness,chow2006comparison}.

For translational effect, AI and DL techniques should be developed in consultation with practicing ophthalmologists and must be validated in “real-world” DR screening programs where fundus images have different qualities (e.g. cataract, poor pupil dilation, poor contrast/focus), in patient samples of different ethnicity (i.e. different fundi pigmentation) and systemic control (poor and good control) \cite{wong2016artificial}.

The benchmark datasets have been used for the evaluation of different methods reviewed here. However, there is a variability in the grading by the human graders and different screening programs differ in local protocols. The benchmark datasets used in the reviewed methods does not follow a standard and as such the methods developed and tested using these datasets might not work in the clinical settings. As the performance of an intelligent computer aided method depends on the dataset that is used to train it, it necessitates the need of the development of new datasets keeping in view the procedures which are adopted in DR screening programs.  Because of the possibility of the variability in grading by human graders while using different classification systems, it is advocated to standardize the use of one classification system for the development of the datasets and to use images that are graded and agreed on by at least 3 different graders. Further, the datasets must be statistically analyzed to determine the accuracy of grading \cite{oke2016use}. In addition, there is a classification introduced by the early treatment of diabetic retinopathy study (ETDRS) \cite{kanski2009clinical} for diabetic maculopathy. It must be taken into account for developing dataset for diabetic maculopathy.

Screening different races is a limitation of DR screening systems. \citet{ting2017development} addressed this issue. However, it is not enough, further research is needed on this issue. From our personal experience, we observed that some DL algorithms developed on western populations were not able to detect some significant lesions in Saudi population.  We noticed that as middle eastern or darkly skinned people may have more melanocytes in their retinal pigmented epithelium (RPE), which forms the most outer layer of the retina. Darker retina obscures some vascular changes as compared to the light colored retina. This limitation was noticed when we started using DL system which was developed using a dataset from light colored retinas. An important challenge to adoption of an algorithm is that it must be validated on larger patient cohorts under different settings and conditions. The performance of a screening software varies with the prevalence of the condition being screened. The prevalence of DR varies and is low in some communities and ethnic groups and higher in others (e.g. Hispanics, African Americans). It is important to understand the performance characteristics in these populations.

Due to the reasons explained above, different datasets of fundus images created for benchmarking DR diagnosis methods are heterogeneous, and a deep-learning-based method gives good performance when trained and tested using the same dataset. For robustness, it is necessary that a deep-learning-based method gives satisfactory performance across different datasets. There are very few methods that have been tested across datasets. As real clinical settings can be forced to match the conditions under which a particular dataset was captured for developing a DR diagnosis method, robust deep-learning methods must be developed to give satisfactory performance in cross-database evaluation, i.e. trained with one dataset and tested with another. 

\section{Conclusion}\label{CON}

Diabetic retinopathy is a complication of diabetes that damages the retina, causing vision problems. Diabetes harms the retinal blood vessels and leads to dangerous consequences, such as blindness. DR is preventable, and to avoid vision loss, early detection is important. Conventional methods for detecting DR biomarkers and lesions are based on hand-engineered features. The advent of deep learning has opened the way to design and develop more robust and accurate methods for detecting and diagnosing different DR complications, and deep learning has been employed to develop many methods for retinal blood vessel segmentation, OD detection and segmentation, detection and classification of different DR lesions, and the detection of referable DR. First, we have given an overview of different DR biomarkers and lesions, different tasks related to DR diagnosis, and the general framework of these tasks. Then we have given an overview of datasets that have been developed for research on DR diagnosis and performance metrics commonly used for evaluation. After that, we give an overview of deep-learning architectures that have been employed for designing DR diagnosis methods. After providing the necessary background, we then reviewed deep-learning-based methods that have been proposed for retinal blood vessels segmentation, OD detection and segmentation, detection of various DR lesions such as EXs, MAs, HMs and referable DR, highlighted their pro and cones and discussed their overall performance and compared them with state-of-the-art traditional methods based on hand--engineered features. In general, the deep-learning approach outperforms the traditional approach based on hand-engineered feature extraction techniques. At the end, we highlighted the gaps and weakness of the existing deep-learning-based DR diagnosis methods and presented potential future directions for research. This review gives a comprehensive view of state-of-the-art deep-learning-based methods related to DR diagnosis and will help researchers to conduct further research on this problem.	 

\par
\fontsize{11}{11}\selectfont 
\bibliographystyle{unsrtnat}
\bibliographystyle{agsm}
\bibliography{ref_}

\begin{thebibliography}{138}
\providecommand{\natexlab}[1]{#1}
\providecommand{\url}[1]{\texttt{#1}}
\expandafter\ifx\csname urlstyle\endcsname\relax
  \providecommand{\doi}[1]{doi: #1}\else
  \providecommand{\doi}{doi: \begingroup \urlstyle{rm}\Url}\fi

\bibitem[Mookiah et~al.(2013)Mookiah, Acharya, Chua, Lim, Ng, and
  Laude]{mookiah2013computer}
Muthu Rama~Krishnan Mookiah, U~Rajendra Acharya, Chua~Kuang Chua, Choo~Min Lim,
  EYK Ng, and Augustinus Laude.
\newblock Computer-aided diagnosis of diabetic retinopathy: A review.
\newblock \emph{Computers in biology and medicine}, 43\penalty0 (12):\penalty0
  2136--2155, 2013.

\bibitem[Faust et~al.(2012)Faust, Acharya, Ng, Ng, and
  Suri]{faust2012algorithms}
Oliver Faust, Rajendra Acharya, Eddie Yin-Kwee Ng, Kwan-Hoong Ng, and Jasjit~S
  Suri.
\newblock Algorithms for the automated detection of diabetic retinopathy using
  digital fundus images: a review.
\newblock \emph{Journal of medical systems}, 36\penalty0 (1):\penalty0
  145--157, 2012.

\bibitem[Joshi and Karule(2018)]{joshi2018review}
Shilpa Joshi and PT~Karule.
\newblock A review on exudates detection methods for diabetic retinopathy.
\newblock \emph{Biomedicine \& Pharmacotherapy}, 97:\penalty0 1454--1460, 2018.

\bibitem[Mansour(2017)]{mansour2017evolutionary}
Romany~F Mansour.
\newblock Evolutionary computing enriched computer-aided diagnosis system for
  diabetic retinopathy: A survey.
\newblock \emph{IEEE reviews in biomedical engineering}, 10:\penalty0 334--349,
  2017.

\bibitem[Almotiri et~al.(2018)Almotiri, Elleithy, and
  Elleithy]{almotiri2018retinal}
Jasem Almotiri, Khaled Elleithy, and Abdelrahman Elleithy.
\newblock Retinal vessels segmentation techniques and algorithms: A survey.
\newblock \emph{Applied Sciences}, 8\penalty0 (2):\penalty0 155, 2018.

\bibitem[Almazroa et~al.(2015)Almazroa, Burman, Raahemifar, and
  Lakshminarayanan]{almazroa2015optic}
Ahmed Almazroa, Ritambhar Burman, Kaamran Raahemifar, and Vasudevan
  Lakshminarayanan.
\newblock Optic disc and optic cup segmentation methodologies for glaucoma
  image detection: a survey.
\newblock \emph{Journal of ophthalmology}, 2015, 2015.

\bibitem[Thakur and Juneja(2018)]{thakur2018survey}
Niharika Thakur and Mamta Juneja.
\newblock Survey on segmentation and classification approaches of optic cup and
  optic disc for diagnosis of glaucoma.
\newblock \emph{Biomedical Signal Processing and Control}, 42:\penalty0
  162--189, 2018.

\bibitem[Group et~al.(1991)]{early1991grading}
Early Treatment Diabetic Retinopathy Study~Research Group et~al.
\newblock Grading diabetic retinopathy from stereoscopic color fundus
  photographs—an extension of the modified airlie house classification: Etdrs
  report number 10.
\newblock \emph{Ophthalmology}, 98\penalty0 (5):\penalty0 786--806, 1991.

\bibitem[Harney(2006)]{harney2006diabetic}
Fiona Harney.
\newblock Diabetic retinopathy.
\newblock \emph{Medicine}, 34\penalty0 (3):\penalty0 95--98, 2006.

\bibitem[McLeod(2005)]{mcleod2005cotton}
D~McLeod.
\newblock Why cotton wool spots should not be regarded as retinal nerve fibre
  layer infarcts.
\newblock \emph{British journal of ophthalmology}, 89\penalty0 (2):\penalty0
  229--237, 2005.

\bibitem[Akram et~al.(2014)Akram, Khalid, Tariq, Khan, and
  Azam]{akram2014detection}
M~Usman Akram, Shehzad Khalid, Anam Tariq, Shoab~A Khan, and Farooque Azam.
\newblock Detection and classification of retinal lesions for grading of
  diabetic retinopathy.
\newblock \emph{Computers in biology and medicine}, 45:\penalty0 161--171,
  2014.

\bibitem[Lee and Cheng(1994)]{lee1994parallel}
Tien-You Lee and HD~Cheng.
\newblock Parallel grading of venous beading on transputer.
\newblock In \emph{Proceedings of 1994 20th Annual Northeast Bioengineering
  Conference}, pages 54--58. IEEE, 1994.

\bibitem[Patz(1980)]{patz1980studies}
Arnall Patz.
\newblock Studies on retinal neovascularization. friedenwald lecture.
\newblock \emph{Investigative ophthalmology \& visual science}, 19\penalty0
  (10):\penalty0 1133--1138, 1980.

\bibitem[dr()]{dr}
Diabetic retinopathy.
\newblock \url{https://www.nhs.uk/conditions/diabetic-retinopathy/stages/}.
\newblock Accessed: 2018-01-08.

\bibitem[Group et~al.(1987)]{early1987treatment}
Early Treatment Diabetic Retinopathy Study~Research Group et~al.
\newblock Treatment techniques and clinical guidelines for photocoagulation of
  diabetic macular edema: Early treatment diabetic retinopathy study report
  number 2.
\newblock \emph{Ophthalmology}, 94\penalty0 (7):\penalty0 761--774, 1987.

\bibitem[Sopharak et~al.(2008)Sopharak, Uyyanonvara, Barman, and
  Williamson]{sopharak2008automatic}
Akara Sopharak, Bunyarit Uyyanonvara, Sarah Barman, and Thomas~H Williamson.
\newblock Automatic detection of diabetic retinopathy exudates from non-dilated
  retinal images using mathematical morphology methods.
\newblock \emph{Computerized medical imaging and graphics}, 32\penalty0
  (8):\penalty0 720--727, 2008.

\bibitem[Jonas et~al.(1988)Jonas, Gusek, and Naumann]{jonas1988optic}
Jost~B Jonas, Gabriele~Ch Gusek, and Gottfried~OH Naumann.
\newblock Optic disk morphometry in high myopia.
\newblock \emph{Graefe's archive for clinical and experimental ophthalmology},
  226\penalty0 (6):\penalty0 587--590, 1988.

\bibitem[Joshi et~al.(2011)Joshi, Sivaswamy, and Krishnadas]{joshi2011optic}
Gopal~Datt Joshi, Jayanthi Sivaswamy, and SR~Krishnadas.
\newblock Optic disk and cup segmentation from monocular color retinal images
  for glaucoma assessment.
\newblock \emph{IEEE transactions on medical imaging}, 30\penalty0
  (6):\penalty0 1192--1205, 2011.

\bibitem[Acharya et~al.(2009)Acharya, Lim, Ng, Chee, and
  Tamura]{acharya2009computer}
Udyavara~R Acharya, Choo~M Lim, E~Yin~Kwee Ng, Caroline Chee, and Toshiyo
  Tamura.
\newblock Computer-based detection of diabetes retinopathy stages using digital
  fundus images.
\newblock \emph{Proceedings of the institution of mechanical engineers, part H:
  journal of engineering in medicine}, 223\penalty0 (5):\penalty0 545--553,
  2009.

\bibitem[Fleming et~al.(2010)Fleming, Goatman, Philip, Williams, Prescott,
  Scotland, McNamee, Leese, Wykes, Sharp, et~al.]{fleming2010role}
Alan~D Fleming, Keith~A Goatman, Sam Philip, Graeme~J Williams, Gordon~J
  Prescott, Graham~S Scotland, Paul McNamee, Graham~P Leese, William~N Wykes,
  Peter~F Sharp, et~al.
\newblock The role of haemorrhage and exudate detection in automated grading of
  diabetic retinopathy.
\newblock \emph{British Journal of Ophthalmology}, 94\penalty0 (6):\penalty0
  706--711, 2010.

\bibitem[dr_()]{dr_grading}
Diabetic retinal screening, grading, monitoring and referral guidance.
\newblock
  \url{https://www.health.govt.nz/publication/diabetic-retinal-screening-grading-monitoring-and-referral-guidance}.
\newblock Accessed: 2019-05-01.

\bibitem[Kanski(2009)]{kanski2009clinical}
Jack~J Kanski.
\newblock \emph{Clinical ophthalmology: a synopsis}.
\newblock Elsevier Health Sciences, 2009.

\bibitem[Zachariah et~al.(2015)Zachariah, Wykes, and
  Yorston]{zachariah2015grading}
Sonia Zachariah, William Wykes, and David Yorston.
\newblock Grading diabetic retinopathy (dr) using the scottish grading
  protocol.
\newblock \emph{Community eye health}, 28\penalty0 (92):\penalty0 72, 2015.

\bibitem[dr-(2015)]{dr-grades}
Diabetic retinopathy (dr): management and referral.
\newblock In \emph{Community Eye Health}, pages 70--71, 2015.

\bibitem[Seoud et~al.(2016)Seoud, Hurtut, Chelbi, Cheriet, and
  Langlois]{seoud2016red}
Lama Seoud, Thomas Hurtut, Jihed Chelbi, Farida Cheriet, and JM~Pierre
  Langlois.
\newblock Red lesion detection using dynamic shape features for diabetic
  retinopathy screening.
\newblock \emph{IEEE transactions on medical imaging}, 35\penalty0
  (4):\penalty0 1116--1126, 2016.

\bibitem[Litjens et~al.(2017)Litjens, Kooi, Bejnordi, Setio, Ciompi,
  Ghafoorian, van~der Laak, van Ginneken, and S{\'a}nchez]{litjens2017survey}
Geert Litjens, Thijs Kooi, Babak~Ehteshami Bejnordi, Arnaud Arindra~Adiyoso
  Setio, Francesco Ciompi, Mohsen Ghafoorian, Jeroen~AWM van~der Laak, Bram van
  Ginneken, and Clara~I S{\'a}nchez.
\newblock A survey on deep learning in medical image analysis.
\newblock \emph{Medical image analysis}, 42:\penalty0 60--88, 2017.

\bibitem[Gondal et~al.(2017)Gondal, K{\"o}hler, Grzeszick, Fink, and
  Hirsch]{Gondaletal17}
W.~Gondal, J.~M. K{\"o}hler, R.~Grzeszick, G.~Fink, and M.~Hirsch.
\newblock Weakly-supervised localization of diabetic retinopathy lesions in
  retinal fundus images.
\newblock In \emph{IEEE International Conference on Image Processing (ICIP
  207)}, 2017.

\bibitem[Li et~al.(2016)Li, Feng, Xie, Liang, Zhang, and Wang]{li2016cross}
Qiaoliang Li, Bowei Feng, LinPei Xie, Ping Liang, Huisheng Zhang, and Tianfu
  Wang.
\newblock A cross-modality learning approach for vessel segmentation in retinal
  images.
\newblock \emph{IEEE transactions on medical imaging}, 35\penalty0
  (1):\penalty0 109--118, 2016.

\bibitem[Arunkumar and Karthigaikumar(2017)]{arunkumar2017multi}
R~Arunkumar and P~Karthigaikumar.
\newblock Multi-retinal disease classification by reduced deep learning
  features.
\newblock \emph{Neural Computing and Applications}, 28\penalty0 (2):\penalty0
  329--334, 2017.

\bibitem[mes()]{messidor}
Messidor dataset.
\newblock \url{http://www.adcis.net/en/Download-Third-Party/Messidor.html}.
\newblock Accessed: 2018-01-08.

\bibitem[opt()]{optha}
E-ophtha.
\newblock \url{http://www.adcis.net/en/Download-Third-Party/E-Ophtha.html}.
\newblock Accessed: 2018-01-08.

\bibitem[kgl()]{kgl}
Kaggle dataset.
\newblock \url{https://www.kaggle.com/c/diabetic-retinopathy-detection/data}.
\newblock Accessed: 2018-01-08.

\bibitem[dri({\natexlab{a}})]{drive}
Drive dataset.
\newblock \url{https://www.isi.uu.nl/Research/Databases/DRIVE/},
  {\natexlab{a}}.
\newblock Accessed: 2018-01-08.

\bibitem[Hoover et~al.(2000)Hoover, Kouznetsova, and
  Goldbaum]{hoover2000locating}
AD~Hoover, Valentina Kouznetsova, and Michael Goldbaum.
\newblock Locating blood vessels in retinal images by piecewise threshold
  probing of a matched filter response.
\newblock \emph{IEEE Transactions on Medical imaging}, 19\penalty0
  (3):\penalty0 203--210, 2000.

\bibitem[K{\"a}lvi{\"a}inen and Uusitalo(2007)]{kalviainen2007diaretdb1}
RVJPH K{\"a}lvi{\"a}inen and H~Uusitalo.
\newblock Diaretdb1 diabetic retinopathy database and evaluation protocol.
\newblock In \emph{Medical Image Understanding and Analysis}, volume 2007,
  page~61. Citeseer, 2007.

\bibitem[dia()]{dia}
Diaretdb1 dataset.
\newblock \url{http://www.it.lut.fi/project/imageret/diaretdb1/}.
\newblock Accessed: 2018-01-08.

\bibitem[cha()]{chase}
Chase dataset.
\newblock \url{http://www.chasestudy.ac.uk/}.
\newblock Accessed: 2018-02-01.

\bibitem[Prentasic et~al.(2013)Prentasic, Loncaric, Vatavuk, Bencic, Subasic,
  Petkovic, Dujmovic, Malenica-Ravlic, Budimlija, and
  Tadic]{prentasic2013diabetic}
Pavle Prentasic, Sven Loncaric, Zoran Vatavuk, Goran Bencic, Marko Subasic,
  Tomislav Petkovic, Lana Dujmovic, Maja Malenica-Ravlic, Nikolina Budimlija,
  and Raseljka Tadic.
\newblock Diabetic retinopathy image database (dridb): a new database for
  diabetic retinopathy screening programs research.
\newblock In \emph{Image and Signal Processing and Analysis (ISPA), 2013 8th
  International Symposium on}, pages 711--716. IEEE, 2013.

\bibitem[Zhang et~al.(2010)Zhang, Yin, Liu, Wong, Tan, Lee, Cheng, and
  Wong]{zhang2010origa}
Zhuo Zhang, Feng~Shou Yin, Jiang Liu, Wing~Kee Wong, Ngan~Meng Tan, Beng~Hai
  Lee, Jun Cheng, and Tien~Yin Wong.
\newblock Origa-light: An online retinal fundus image database for glaucoma
  analysis and research.
\newblock In \emph{Engineering in Medicine and Biology Society (EMBC), 2010
  Annual International Conference of the IEEE}, pages 3065--3068. IEEE, 2010.

\bibitem[Sng et~al.(2012)Sng, Foo, Cheng, Allen, He, Krishnaswamy, Nongpiur,
  Friedman, Wong, and Aung]{sng2012determinants}
Chelvin~C Sng, Li-Lian Foo, Ching-Yu Cheng, John~C Allen, Mingguang He, Gita
  Krishnaswamy, Monisha~E Nongpiur, David~S Friedman, Tien~Y Wong, and Tin
  Aung.
\newblock Determinants of anterior chamber depth: the singapore chinese eye
  study.
\newblock \emph{Ophthalmology}, 119\penalty0 (6):\penalty0 1143--1150, 2012.

\bibitem[nih()]{nih}
Nih areds dataset.
\newblock
  \url{https://www.nih.gov/news-events/news-releases/nih-adds-first-images-major-research-database}.
\newblock Accessed: 2018-02-01.

\bibitem[Al-Diri et~al.(2008)Al-Diri, Hunter, Steel, Habib, Hudaib, and
  Berry]{al2008reference}
Bashir Al-Diri, Andrew Hunter, David Steel, Maged Habib, Taghread Hudaib, and
  Simon Berry.
\newblock A reference data set for retinal vessel profiles.
\newblock In \emph{Engineering in Medicine and Biology Society, 2008. EMBS
  2008. 30th Annual International Conference of the IEEE}, pages 2262--2265.
  IEEE, 2008.

\bibitem[eye()]{eyepacs}
Eyepacs dataset.
\newblock \url{http://www.eyepacs.com/eyepacssystem/}.
\newblock Accessed: 2018-03-01.

\bibitem[Fumero et~al.(2011)Fumero, Alay{\'o}n, Sanchez, Sigut, and
  Gonzalez-Hernandez]{fumero2011rim}
Francisco Fumero, Silvia Alay{\'o}n, JL~Sanchez, J~Sigut, and
  M~Gonzalez-Hernandez.
\newblock Rim-one: An open retinal image database for optic nerve evaluation.
\newblock In \emph{Computer-Based Medical Systems (CBMS), 2011 24th
  International Symposium on}, pages 1--6. IEEE, 2011.

\bibitem[Sivaswamy et~al.(2014)Sivaswamy, Krishnadas, Joshi, Jain, and
  Tabish]{sivaswamy2014drishti}
Jayanthi Sivaswamy, SR~Krishnadas, Gopal~Datt Joshi, Madhulika Jain, and
  A~Ujjwaft~Syed Tabish.
\newblock Drishti-gs: Retinal image dataset for optic nerve head (onh)
  segmentation.
\newblock In \emph{Biomedical Imaging (ISBI), 2014 IEEE 11th International
  Symposium on}, pages 53--56. IEEE, 2014.

\bibitem[ari()]{aria}
Aria dataset.
\newblock \url{http://www.eyecharity.com/aria_online.html}.
\newblock Accessed: 2018-02-28.

\bibitem[dri({\natexlab{b}})]{drion}
drion dataset.
\newblock \url{http://www.ia.uned.es/~ejcarmona/DRIONS-DB.html},
  {\natexlab{b}}.
\newblock Accessed: 2018-04-30.

\bibitem[see()]{seed}
Seed-db.
\newblock
  \url{https://www.seri.com.sg/key-programmes/singapore-epidemiology-of-eye-diseases-seed/}.
\newblock Assessed on 2018-08-08.

\bibitem[Decenci{\`e}re et~al.(2014)Decenci{\`e}re, Zhang, Cazuguel, Lay,
  Cochener, Trone, Gain, Ordonez, Massin, Erginay,
  et~al.]{decenciere2014feedback}
Etienne Decenci{\`e}re, Xiwei Zhang, Guy Cazuguel, Bruno Lay, B{\'e}atrice
  Cochener, Caroline Trone, Philippe Gain, Richard Ordonez, Pascale Massin, Ali
  Erginay, et~al.
\newblock Feedback on a publicly distributed image database: the messidor
  database.
\newblock \emph{Image Analysis \& Stereology}, 33\penalty0 (3):\penalty0
  231--234, 2014.

\bibitem[Mitchell(2004)]{mitchell2004role}
Tom~M Mitchell.
\newblock The role of unlabeled data in supervised learning.
\newblock In \emph{Language, Knowledge, and Representation}, pages 103--111.
  Springer, 2004.

\bibitem[Shankaranarayana et~al.(2017)Shankaranarayana, Ram, Mitra, and
  Sivaprakasam]{shankaranarayana2017joint}
Sharath~M Shankaranarayana, Keerthi Ram, Kaushik Mitra, and Mohanasankar
  Sivaprakasam.
\newblock Joint optic disc and cup segmentation using fully convolutional and
  adversarial networks.
\newblock In \emph{Fetal, Infant and Ophthalmic Medical Image Analysis}, pages
  168--176. Springer, 2017.

\bibitem[Srivastava et~al.(2015)Srivastava, Cheng, Wong, and
  Liu]{srivastava2015using}
Ruchir Srivastava, Jun Cheng, Damon~WK Wong, and Jiang Liu.
\newblock Using deep learning for robustness to parapapillary atrophy in optic
  disc segmentation.
\newblock In \emph{Biomedical Imaging (ISBI), 2015 IEEE 12th International
  Symposium on}, pages 768--771. IEEE, 2015.

\bibitem[Zou et~al.(2004)Zou, Warfield, Bharatha, Tempany, Kaus, Haker, Wells,
  Jolesz, and Kikinis]{zou2004statistical}
Kelly~H Zou, Simon~K Warfield, Aditya Bharatha, Clare~MC Tempany, Michael~R
  Kaus, Steven~J Haker, William~M Wells, Ferenc~A Jolesz, and Ron Kikinis.
\newblock Statistical validation of image segmentation quality based on a
  spatial overlap index1: scientific reports.
\newblock \emph{Academic radiology}, 11\penalty0 (2):\penalty0 178--189, 2004.

\bibitem[Zhang et~al.(2015)Zhang, Feng, Xiao, He, and
  Zhu]{zhang2015segmentation}
Xueliang Zhang, Xuezhi Feng, Pengfeng Xiao, Guangjun He, and Liujun Zhu.
\newblock Segmentation quality evaluation using region-based precision and
  recall measures for remote sensing images.
\newblock \emph{ISPRS Journal of Photogrammetry and Remote Sensing},
  102:\penalty0 73--84, 2015.

\bibitem[LeCun et~al.(1998)LeCun, Bottou, Bengio, and
  Haffner]{lecun1998gradient}
Yann LeCun, L{\'e}on Bottou, Yoshua Bengio, and Patrick Haffner.
\newblock Gradient-based learning applied to document recognition.
\newblock \emph{Proceedings of the IEEE}, 86\penalty0 (11):\penalty0
  2278--2324, 1998.

\bibitem[Krizhevsky et~al.(2012)Krizhevsky, Sutskever, and
  Hinton]{krizhevsky2012imagenet}
Alex Krizhevsky, Ilya Sutskever, and Geoffrey~E Hinton.
\newblock Imagenet classification with deep convolutional neural networks.
\newblock In \emph{Advances in neural information processing systems}, pages
  1097--1105, 2012.

\bibitem[Simonyan and Zisserman(2014)]{simonyan2014very}
Karen Simonyan and Andrew Zisserman.
\newblock Very deep convolutional networks for large-scale image recognition.
\newblock \emph{arXiv preprint arXiv:1409.1556}, 2014.

\bibitem[Szegedy et~al.(2015)Szegedy, Liu, Jia, Sermanet, Reed, Anguelov,
  Erhan, Vanhoucke, Rabinovich, et~al.]{szegedy2015going}
Christian Szegedy, Wei Liu, Yangqing Jia, Pierre Sermanet, Scott Reed, Dragomir
  Anguelov, Dumitru Erhan, Vincent Vanhoucke, Andrew Rabinovich, et~al.
\newblock Going deeper with convolutions.
\newblock Cvpr, 2015.

\bibitem[He et~al.(2016)He, Zhang, Ren, and Sun]{he2016deep}
Kaiming He, Xiangyu Zhang, Shaoqing Ren, and Jian Sun.
\newblock Deep residual learning for image recognition.
\newblock In \emph{Proceedings of the IEEE conference on computer vision and
  pattern recognition}, pages 770--778, 2016.

\bibitem[Tajbakhsh et~al.(2016)Tajbakhsh, Shin, Gurudu, Hurst, Kendall, Gotway,
  and Liang]{tajbakhsh2016convolutional}
Nima Tajbakhsh, Jae~Y Shin, Suryakanth~R Gurudu, R~Todd Hurst, Christopher~B
  Kendall, Michael~B Gotway, and Jianming Liang.
\newblock Convolutional neural networks for medical image analysis: Full
  training or fine tuning?
\newblock \emph{IEEE transactions on medical imaging}, 35\penalty0
  (5):\penalty0 1299--1312, 2016.

\bibitem[Erhan et~al.(2010)Erhan, Bengio, Courville, Manzagol, Vincent, and
  Bengio]{erhan2010does}
Dumitru Erhan, Yoshua Bengio, Aaron Courville, Pierre-Antoine Manzagol, Pascal
  Vincent, and Samy Bengio.
\newblock Why does unsupervised pre-training help deep learning?
\newblock \emph{Journal of Machine Learning Research}, 11\penalty0
  (Feb):\penalty0 625--660, 2010.

\bibitem[Long et~al.(2015)Long, Shelhamer, and Darrell]{long2015fully}
Jonathan Long, Evan Shelhamer, and Trevor Darrell.
\newblock Fully convolutional networks for semantic segmentation.
\newblock In \emph{Proceedings of the IEEE conference on computer vision and
  pattern recognition}, pages 3431--3440, 2015.

\bibitem[Hinton and Salakhutdinov(2006)]{hinton2006reducing}
Geoffrey~E Hinton and Ruslan~R Salakhutdinov.
\newblock Reducing the dimensionality of data with neural networks.
\newblock \emph{science}, 313\penalty0 (5786):\penalty0 504--507, 2006.

\bibitem[Liou et~al.(2014)Liou, Cheng, Liou, and Liou]{liou2014autoencoder}
Cheng-Yuan Liou, Wei-Chen Cheng, Jiun-Wei Liou, and Daw-Ran Liou.
\newblock Autoencoder for words.
\newblock \emph{Neurocomputing}, 139:\penalty0 84--96, 2014.

\bibitem[Maji et~al.(2015)Maji, Santara, Ghosh, Sheet, and Mitra]{maji2015deep}
Debapriya Maji, Anirban Santara, Sambuddha Ghosh, Debdoot Sheet, and Pabitra
  Mitra.
\newblock Deep neural network and random forest hybrid architecture for
  learning to detect retinal vessels in fundus images.
\newblock In \emph{Engineering in Medicine and Biology Society (EMBC), 2015
  37th Annual International Conference of the IEEE}, pages 3029--3032. IEEE,
  2015.

\bibitem[Mikolov et~al.(2010)Mikolov, Karafi{\'a}t, Burget, {\v{C}}ernock{\`y},
  and Khudanpur]{mikolov2010recurrent}
Tom{\'a}{\v{s}} Mikolov, Martin Karafi{\'a}t, Luk{\'a}{\v{s}} Burget, Jan
  {\v{C}}ernock{\`y}, and Sanjeev Khudanpur.
\newblock Recurrent neural network based language model.
\newblock In \emph{Eleventh Annual Conference of the International Speech
  Communication Association}, 2010.

\bibitem[Vinyals et~al.(2017)Vinyals, Toshev, Bengio, and
  Erhan]{vinyals2017show}
Oriol Vinyals, Alexander Toshev, Samy Bengio, and Dumitru Erhan.
\newblock Show and tell: Lessons learned from the 2015 mscoco image captioning
  challenge.
\newblock \emph{IEEE transactions on pattern analysis and machine
  intelligence}, 39\penalty0 (4):\penalty0 652--663, 2017.

\bibitem[Maji et~al.(2016)Maji, Santara, Mitra, and Sheet]{maji2016ensemble}
Debapriya Maji, Anirban Santara, Pabitra Mitra, and Debdoot Sheet.
\newblock Ensemble of deep convolutional neural networks for learning to detect
  retinal vessels in fundus images.
\newblock \emph{arXiv preprint arXiv:1603.04833}, 2016.

\bibitem[Liskowski and Krawiec(2016)]{liskowski2016segmenting}
Pawe{\l} Liskowski and Krzysztof Krawiec.
\newblock Segmenting retinal blood vessels with deep neural networks.
\newblock \emph{IEEE transactions on medical imaging}, 35\penalty0
  (11):\penalty0 2369--2380, 2016.

\bibitem[Maninis et~al.(2016)Maninis, Pont-Tuset, Arbel{\'a}ez, and
  Van~Gool]{maninis2016deep}
Kevis-Kokitsi Maninis, Jordi Pont-Tuset, Pablo Arbel{\'a}ez, and Luc Van~Gool.
\newblock Deep retinal image understanding.
\newblock In \emph{International Conference on Medical Image Computing and
  Computer-Assisted Intervention}, pages 140--148. Springer, 2016.

\bibitem[Wu et~al.(2016)Wu, Xu, Gao, Buty, and Mollura]{wu2016deep}
Aaron Wu, Ziyue Xu, Mingchen Gao, Mario Buty, and Daniel~J Mollura.
\newblock Deep vessel tracking: A generalized probabilistic approach via deep
  learning.
\newblock In \emph{Biomedical Imaging (ISBI), 2016 IEEE 13th International
  Symposium on}, pages 1363--1367. IEEE, 2016.

\bibitem[Dasgupta and Singh(2017)]{dasgupta2017fully}
Avijit Dasgupta and Sonam Singh.
\newblock A fully convolutional neural network based structured prediction
  approach towards the retinal vessel segmentation.
\newblock In \emph{Biomedical Imaging (ISBI 2017), 2017 IEEE 14th International
  Symposium on}, pages 248--251. IEEE, 2017.

\bibitem[Tan et~al.(2017)Tan, Acharya, Bhandary, Chua, and
  Sivaprasad]{tan2017segmentation}
Jen~Hong Tan, U~Rajendra Acharya, Sulatha~V Bhandary, Kuang~Chua Chua, and
  Sobha Sivaprasad.
\newblock Segmentation of optic disc, fovea and retinal vasculature using a
  single convolutional neural network.
\newblock \emph{Journal of Computational Science}, 20:\penalty0 70--79, 2017.

\bibitem[Fu et~al.(2016{\natexlab{a}})Fu, Xu, Wong, and Liu]{fu2016retinal}
Huazhu Fu, Yanwu Xu, Damon Wing~Kee Wong, and Jiang Liu.
\newblock Retinal vessel segmentation via deep learning network and
  fully-connected conditional random fields.
\newblock In \emph{Biomedical Imaging (ISBI), 2016 IEEE 13th International
  Symposium on}, pages 698--701. IEEE, 2016{\natexlab{a}}.

\bibitem[Mo and Zhang(2017)]{mo2017multi}
Juan Mo and Lei Zhang.
\newblock Multi-level deep supervised networks for retinal vessel segmentation.
\newblock \emph{International journal of computer assisted radiology and
  surgery}, 12\penalty0 (12):\penalty0 2181--2193, 2017.

\bibitem[Roy and Sheet(2015)]{roy2015dasa}
Abhijit~Guha Roy and Debdoot Sheet.
\newblock Dasa: Domain adaptation in stacked autoencoders using systematic
  dropout.
\newblock In \emph{Pattern Recognition (ACPR), 2015 3rd IAPR Asian Conference
  on}, pages 735--739. IEEE, 2015.

\bibitem[Lahiri et~al.(2016)Lahiri, Roy, Sheet, and Biswas]{lahiri2016deep}
Avisek Lahiri, Abhijit~Guha Roy, Debdoot Sheet, and Prabir~Kumar Biswas.
\newblock Deep neural ensemble for retinal vessel segmentation in fundus images
  towards achieving label-free angiography.
\newblock In \emph{Engineering in Medicine and Biology Society (EMBC), 2016
  IEEE 38th Annual International Conference of the}, pages 1340--1343. IEEE,
  2016.

\bibitem[Fu et~al.(2016{\natexlab{b}})Fu, Xu, Lin, Wong, and
  Liu]{fu2016deepvessel}
Huazhu Fu, Yanwu Xu, Stephen Lin, Damon Wing~Kee Wong, and Jiang Liu.
\newblock Deepvessel: Retinal vessel segmentation via deep learning and
  conditional random field.
\newblock In \emph{International Conference on Medical Image Computing and
  Computer-Assisted Intervention}, pages 132--139. Springer,
  2016{\natexlab{b}}.

\bibitem[Lim et~al.(2015)Lim, Cheng, Hsu, and Lee]{lim2015integrated}
Gilbert Lim, Yuan Cheng, Wynne Hsu, and Mong~Li Lee.
\newblock Integrated optic disc and cup segmentation with deep learning.
\newblock pages 162--169, 2015.

\bibitem[Guo et~al.(2016)Guo, Zou, Chen, He, Liu, and Zhao]{guo2016optic}
Yundi Guo, Beiji Zou, Zailiang Chen, Qi~He, Qing Liu, and Rongchang Zhao.
\newblock Optic cup segmentation using large pixel patch based cnns.
\newblock 2016.

\bibitem[Sevastopolsky(2017)]{sevastopolsky2017optic}
Artem Sevastopolsky.
\newblock Optic disc and cup segmentation methods for glaucoma detection with
  modification of u-net convolutional neural network.
\newblock \emph{Pattern Recognition and Image Analysis}, 27\penalty0
  (3):\penalty0 618--624, 2017.

\bibitem[Ronneberger et~al.(2015)Ronneberger, Fischer, and
  Brox]{ronneberger2015u}
Olaf Ronneberger, Philipp Fischer, and Thomas Brox.
\newblock U-net: Convolutional networks for biomedical image segmentation.
\newblock In \emph{International Conference on Medical image computing and
  computer-assisted intervention}, pages 234--241. Springer, 2015.

\bibitem[Zilly et~al.(2015)Zilly, Buhmann, and Mahapatra]{zilly2015boosting}
Julian~G Zilly, Joachim~M Buhmann, and Dwarikanath Mahapatra.
\newblock Boosting convolutional filters with entropy sampling for optic cup
  and disc image segmentation from fundus images.
\newblock pages 136--143, 2015.

\bibitem[Zilly et~al.(2017)Zilly, Buhmann, and Mahapatra]{zilly2017glaucoma}
Julian Zilly, Joachim~M Buhmann, and Dwarikanath Mahapatra.
\newblock Glaucoma detection using entropy sampling and ensemble learning for
  automatic optic cup and disc segmentation.
\newblock \emph{Computerized Medical Imaging and Graphics}, 55:\penalty0
  28--41, 2017.

\bibitem[Zhang et~al.(2018)Zhang, Zhu, Zhao, Shi, and Chen]{zhang2018automatic}
Defeng Zhang, Weifang Zhu, Heming Zhao, Fei Shi, and Xinjian Chen.
\newblock Automatic localization and segmentation of optical disk based on
  faster r-cnn and level set in fundus image.
\newblock In \emph{Medical Imaging 2018: Image Processing}, volume 10574, page
  105741U. International Society for Optics and Photonics, 2018.

\bibitem[Fu et~al.(2018)Fu, Cheng, Xu, Wong, Liu, and Cao]{fu2018joint}
Huazhu Fu, Jun Cheng, Yanwu Xu, Damon Wing~Kee Wong, Jiang Liu, and Xiaochun
  Cao.
\newblock Joint optic disc and cup segmentation based on multi-label deep
  network and polar transformation.
\newblock \emph{arXiv preprint arXiv:1801.00926}, 2018.

\bibitem[Niu et~al.(2017)Niu, Xu, Wan, Cheng, and Liu]{niu2017automatic}
Di~Niu, Peiyuan Xu, Cheng Wan, Jun Cheng, and Jiang Liu.
\newblock Automatic localization of optic disc based on deep learning in fundus
  images.
\newblock In \emph{Signal and Image Processing (ICSIP), 2017 IEEE 2nd
  International Conference on}, pages 208--212. IEEE, 2017.

\bibitem[Alghamdi et~al.(2016)Alghamdi, Tang, Waheeb, and
  Peto]{alghamdi2016automatic}
Hanan~S Alghamdi, Hongying~Lilian Tang, Saad~A Waheeb, and Tunde Peto.
\newblock Automatic optic disc abnormality detection in fundus images: a deep
  learning approach.
\newblock 2016.

\bibitem[Xu et~al.(2017)Xu, Wan, Cheng, Niu, and Liu]{xu2017optic}
Peiyuan Xu, Cheng Wan, Jun Cheng, Di~Niu, and Jiang Liu.
\newblock Optic disc detection via deep learning in fundus images.
\newblock In \emph{Fetal, Infant and Ophthalmic Medical Image Analysis}, pages
  134--141. Springer, 2017.

\bibitem[Foong et~al.(2007)Foong, Saw, Loo, Shen, Loon, Rosman, Aung, Tan, Tai,
  and Wong]{foong2007rationale}
Athena~WP Foong, Seang-Mei Saw, Jing-Liang Loo, Sunny Shen, Seng-Chee Loon,
  Mohamad Rosman, Tin Aung, Donald~TH Tan, E~Shyong Tai, and Tien~Y Wong.
\newblock Rationale and methodology for a population-based study of eye
  diseases in malay people: The singapore malay eye study (simes).
\newblock \emph{Ophthalmic epidemiology}, 14\penalty0 (1):\penalty0 25--35,
  2007.

\bibitem[Abr{\`a}moff et~al.(2016)Abr{\`a}moff, Lou, Erginay, Clarida, Amelon,
  Folk, and Niemeijer]{abramoff2016improved}
Michael~David Abr{\`a}moff, Yiyue Lou, Ali Erginay, Warren Clarida, Ryan
  Amelon, James~C Folk, and Meindert Niemeijer.
\newblock Improved automated detection of diabetic retinopathy on a publicly
  available dataset through integration of deep learning.
\newblock \emph{Investigative ophthalmology \& visual science}, 57\penalty0
  (13):\penalty0 5200--5206, 2016.

\bibitem[Perdomo et~al.(2016)Perdomo, Otalora, Rodr{\'\i}guez, Arevalo, and
  Gonz{\'a}lez]{perdomo2016novel}
Oscar Perdomo, Sebastian Otalora, Francisco Rodr{\'\i}guez, John Arevalo, and
  Fabio~A Gonz{\'a}lez.
\newblock A novel machine learning model based on exudate localization to
  detect diabetic macular edema.
\newblock 2016.

\bibitem[Burlina et~al.(2016)Burlina, Freund, Joshi, Wolfson, and
  Bressler]{burlina2016detection}
Philippe Burlina, David~E Freund, Neil Joshi, Y~Wolfson, and Neil~M Bressler.
\newblock Detection of age-related macular degeneration via deep learning.
\newblock In \emph{Biomedical Imaging (ISBI), 2016 IEEE 13th International
  Symposium on}, pages 184--188. IEEE, 2016.

\bibitem[Al-Bander et~al.(2016)Al-Bander, Al-Nuaimy, Al-Taee, Williams, and
  Zheng]{al2016diabetic}
Baidaa Al-Bander, Waleed Al-Nuaimy, Majid~A Al-Taee, Bryan~M Williams, and
  Yalin Zheng.
\newblock Diabetic macular edema grading based on deep neural networks.
\newblock 2016.

\bibitem[Ting et~al.(2017)Ting, Cheung, Lim, Tan, Quang, Gan, Hamzah,
  Garcia-Franco, San~Yeo, Lee, et~al.]{ting2017development}
Daniel Shu~Wei Ting, Carol Yim-Lui Cheung, Gilbert Lim, Gavin Siew~Wei Tan,
  Nguyen~D Quang, Alfred Gan, Haslina Hamzah, Renata Garcia-Franco, Ian~Yew
  San~Yeo, Shu~Yen Lee, et~al.
\newblock Development and validation of a deep learning system for diabetic
  retinopathy and related eye diseases using retinal images from multiethnic
  populations with diabetes.
\newblock \emph{Jama}, 318\penalty0 (22):\penalty0 2211--2223, 2017.

\bibitem[Mo et~al.(2018)Mo, Zhang, and Feng]{mo2018exudate}
Juan Mo, Lei Zhang, and Yangqin Feng.
\newblock Exudate-based diabetic macular edema recognition in retinal images
  using cascaded deep residual networks.
\newblock \emph{Neurocomputing}, 290:\penalty0 161--171, 2018.

\bibitem[hei()]{hei-med}
Hei-med dataset.
\newblock \url{http://www.genenetwork.org/dbdoc/Eye_M2_0908_R.html}.
\newblock Accessed: 2018-11-02.

\bibitem[Prenta{\v{s}}i{\'c} and
  Lon{\v{c}}ari{\'c}(2016)]{prentavsic2016detection}
Pavle Prenta{\v{s}}i{\'c} and Sven Lon{\v{c}}ari{\'c}.
\newblock Detection of exudates in fundus photographs using deep neural
  networks and anatomical landmark detection fusion.
\newblock \emph{Computer methods and programs in biomedicine}, 137:\penalty0
  281--292, 2016.

\bibitem[Perdomo et~al.(2017)Perdomo, Arevalo, and
  Gonz{\'a}lez]{perdomo2017convolutional}
Oscar Perdomo, John Arevalo, and Fabio~A Gonz{\'a}lez.
\newblock Convolutional network to detect exudates in eye fundus images of
  diabetic subjects.
\newblock In \emph{12th International Symposium on Medical Information
  Processing and Analysis}, volume 10160, page 101600T. International Society
  for Optics and Photonics, 2017.

\bibitem[oO()]{oO}
o o cnn solution.
\newblock
  \url{https://www.kaggle.com/c/diabetic-retinopathydetection/discussion/15617}.
\newblock Accessed: 2017-01-16.

\bibitem[Zhou et~al.(2016)Zhou, Khosla, Lapedriza, Oliva, and
  Torralba]{zhou2016learning}
Bolei Zhou, Aditya Khosla, Agata Lapedriza, Aude Oliva, and Antonio Torralba.
\newblock Learning deep features for discriminative localization.
\newblock In \emph{Proceedings of the IEEE Conference on Computer Vision and
  Pattern Recognition}, pages 2921--2929, 2016.

\bibitem[Quellec et~al.(2017)Quellec, Charri{\`e}re, Boudi, Cochener, and
  Lamard]{quellec2017deep}
Gwenol{\'e} Quellec, Katia Charri{\`e}re, Yassine Boudi, B{\'e}atrice Cochener,
  and Mathieu Lamard.
\newblock Deep image mining for diabetic retinopathy screening.
\newblock \emph{Medical Image Analysis}, 39:\penalty0 178--193, 2017.

\bibitem[Khojasteh et~al.(2018)Khojasteh, J{\'u}nior, Carvalho, Rezende,
  Aliahmad, Papa, and Kumar]{khojasteh2018exudate}
Parham Khojasteh, Leandro Aparecido~Passos J{\'u}nior, Tiago Carvalho, Edmar
  Rezende, Behzad Aliahmad, Jo{\~a}o~Paulo Papa, and Dinesh~Kant Kumar.
\newblock Exudate detection in fundus images using deeply-learnable features.
\newblock \emph{Computers in biology and medicine}, 2018.

\bibitem[Haloi(2015)]{haloi2015improved}
Mrinal Haloi.
\newblock Improved microaneurysm detection using deep neural networks.
\newblock \emph{arXiv preprint arXiv:1505.04424}, 2015.

\bibitem[van Grinsven et~al.(2016)van Grinsven, van Ginneken, Hoyng, Theelen,
  and S{\'a}nchez]{van2016fast}
Mark~JJP van Grinsven, Bram van Ginneken, Carel~B Hoyng, Thomas Theelen, and
  Clara~I S{\'a}nchez.
\newblock Fast convolutional neural network training using selective data
  sampling: Application to hemorrhage detection in color fundus images.
\newblock \emph{IEEE transactions on medical imaging}, 35\penalty0
  (5):\penalty0 1273--1284, 2016.

\bibitem[Orlando et~al.(2018)Orlando, Prokofyeva, del Fresno, and
  Blaschko]{orlando2018ensemble}
Jos{\'e}~Ignacio Orlando, Elena Prokofyeva, Mariana del Fresno, and Matthew~B
  Blaschko.
\newblock An ensemble deep learning based approach for red lesion detection in
  fundus images.
\newblock \emph{Computer methods and programs in biomedicine}, 153:\penalty0
  115--127, 2018.

\bibitem[Shan and Li(2016)]{shan2016deep}
Juan Shan and Lin Li.
\newblock A deep learning method for microaneurysm detection in fundus images.
\newblock In \emph{Connected Health: Applications, Systems and Engineering
  Technologies (CHASE), 2016 IEEE First International Conference on}, pages
  357--358. IEEE, 2016.

\bibitem[Gulshan et~al.(2016)Gulshan, Peng, Coram, Stumpe, Wu, Narayanaswamy,
  Venugopalan, Widner, Madams, Cuadros, et~al.]{gulshan2016development}
Varun Gulshan, Lily Peng, Marc Coram, Martin~C Stumpe, Derek Wu, Arunachalam
  Narayanaswamy, Subhashini Venugopalan, Kasumi Widner, Tom Madams, Jorge
  Cuadros, et~al.
\newblock Development and validation of a deep learning algorithm for detection
  of diabetic retinopathy in retinal fundus photographs.
\newblock \emph{Jama}, 316\penalty0 (22):\penalty0 2402--2410, 2016.

\bibitem[Colas et~al.(2016)Colas, Besse, Orgogozo, Schmauch, Meric, and
  Besse]{colas2016deep}
E~Colas, A~Besse, A~Orgogozo, B~Schmauch, N~Meric, and E~Besse.
\newblock Deep learning approach for diabetic retinopathy screening.
\newblock \emph{Acta Ophthalmologica}, 94\penalty0 (S256), 2016.

\bibitem[Costa and Campilho(2017)]{costa2017convolutional}
Pedro Costa and Aur{\'e}lio Campilho.
\newblock Convolutional bag of words for diabetic retinopathy detection from
  eye fundus images.
\newblock \emph{IPSJ Transactions on Computer Vision and Applications},
  9\penalty0 (1):\penalty0 10, 2017.

\bibitem[Pires et~al.(2014)Pires, Jelinek, Wainer, Valle, and
  Rocha]{pires2014advancing}
Ramon Pires, Herbert~F Jelinek, Jacques Wainer, Eduardo Valle, and Anderson
  Rocha.
\newblock Advancing bag-of-visual-words representations for lesion
  classification in retinal images.
\newblock \emph{PloS one}, 9\penalty0 (6):\penalty0 e96814, 2014.

\bibitem[Pratt et~al.(2016)Pratt, Coenen, Broadbent, Harding, and
  Zheng]{pratt2016convolutional}
Harry Pratt, Frans Coenen, Deborah~M Broadbent, Simon~P Harding, and Yalin
  Zheng.
\newblock Convolutional neural networks for diabetic retinopathy.
\newblock \emph{Procedia Computer Science}, 90:\penalty0 200--205, 2016.

\bibitem[Gargeya and Leng(2017)]{gargeya2017automated}
Rishab Gargeya and Theodore Leng.
\newblock Automated identification of diabetic retinopathy using deep learning.
\newblock \emph{Ophthalmology}, 124\penalty0 (7):\penalty0 962--969, 2017.

\bibitem[Wang et~al.(2017)Wang, Yin, Shi, Fang, Li, and Wang]{wang2017zoom}
Zhe Wang, Yanxin Yin, Jianping Shi, Wei Fang, Hongsheng Li, and Xiaogang Wang.
\newblock Zoom-in-net: Deep mining lesions for diabetic retinopathy detection.
\newblock In \emph{International Conference on Medical Image Computing and
  Computer-Assisted Intervention}, pages 267--275. Springer, 2017.

\bibitem[Mansour(2018)]{mansour2018deep}
Romany~F Mansour.
\newblock Deep-learning-based automatic computer-aided diagnosis system for
  diabetic retinopathy.
\newblock \emph{Biomedical Engineering Letters}, 8\penalty0 (1):\penalty0
  41--57, 2018.

\bibitem[Chen et~al.(2018)Chen, Wu, Wong, and Lee]{chen2018diabetic}
Yi-Wei Chen, Tung-Yu Wu, Wing-Hung Wong, and Chen-Yi Lee.
\newblock Diabetic retinopathy detection based on deep convolutional neural
  networks.
\newblock In \emph{2018 IEEE International Conference on Acoustics, Speech and
  Signal Processing (ICASSP)}, pages 1030--1034. IEEE, 2018.

\bibitem[Srinidhi et~al.(2017)Srinidhi, Aparna, and Rajan]{srinidhi2017recent}
Chetan~L Srinidhi, P~Aparna, and Jeny Rajan.
\newblock Recent advancements in retinal vessel segmentation.
\newblock \emph{Journal of medical systems}, 41\penalty0 (4):\penalty0 70,
  2017.

\bibitem[Villalobos-Castaldi et~al.(2010)Villalobos-Castaldi,
  Felipe-River{\'o}n, and S{\'a}nchez-Fern{\'a}ndez]{villalobos2010fast}
Fabiola~M Villalobos-Castaldi, Edgardo~M Felipe-River{\'o}n, and Luis~P
  S{\'a}nchez-Fern{\'a}ndez.
\newblock A fast, efficient and automated method to extract vessels from fundus
  images.
\newblock \emph{Journal of Visualization}, 13\penalty0 (3):\penalty0 263--270,
  2010.

\bibitem[Condurache and Mertins(2012)]{condurache2012segmentation}
Alexandru~Paul Condurache and Alfred Mertins.
\newblock Segmentation of retinal vessels with a hysteresis
  binary-classification paradigm.
\newblock \emph{Computerized Medical Imaging and Graphics}, 36\penalty0
  (4):\penalty0 325--335, 2012.

\bibitem[Aquino et~al.(2010)Aquino, Geg{\'u}ndez-Arias, and
  Mar{\'\i}n]{aquino2010detecting}
Arturo Aquino, Manuel~Emilio Geg{\'u}ndez-Arias, and Diego Mar{\'\i}n.
\newblock Detecting the optic disc boundary in digital fundus images using
  morphological, edge detection, and feature extraction techniques.
\newblock \emph{IEEE transactions on medical imaging}, 29\penalty0
  (11):\penalty0 1860--1869, 2010.

\bibitem[Zhang et~al.(2012)Zhang, Yi, Shang, and Peng]{zhang2012optic}
Dongbo Zhang, Yao Yi, Xingyu Shang, and Yinghui Peng.
\newblock Optic disc localization by projection with vessel distribution and
  appearance characteristics.
\newblock In \emph{Pattern Recognition (ICPR), 2012 21st International
  Conference on}, pages 3176--3179. IEEE, 2012.

\bibitem[Sinha and Babu(2012)]{sinha2012optic}
Neelam Sinha and R~Venkatesh Babu.
\newblock Optic disk localization using l 1 minimization.
\newblock In \emph{Image Processing (ICIP), 2012 19th IEEE International
  Conference on}, pages 2829--2832. IEEE, 2012.

\bibitem[Tjandrasa et~al.(2012)Tjandrasa, Wijayanti, and
  Suciati]{tjandrasa2012optic}
Handayani Tjandrasa, Ari Wijayanti, and Nanik Suciati.
\newblock Optic nerve head segmentation using hough transform and active
  contours.
\newblock \emph{Indonesian Journal of Electrical Engineering and Computer
  Science}, 10\penalty0 (3):\penalty0 531--536, 2012.

\bibitem[Massey et~al.(2009)Massey, Hunter, Lowell, and
  Steel]{massey2009robust}
EM~Massey, Andrew Hunter, James Lowell, and DH~Steel.
\newblock A robust lesion boundary segmentation algorithm using level set
  methods.
\newblock In \emph{World Congress on Medical Physics and Biomedical
  Engineering, September 7-12, 2009, Munich, Germany}, pages 304--307.
  Springer, 2009.

\bibitem[Antal and Hajdu(2012)]{antal2012ensemble}
Balint Antal and Andras Hajdu.
\newblock An ensemble-based system for microaneurysm detection and diabetic
  retinopathy grading.
\newblock \emph{IEEE transactions on biomedical engineering}, 59\penalty0
  (6):\penalty0 1720--1726, 2012.

\bibitem[Neff et~al.(2017)Neff, Payer, {\v{S}}tern, and
  Urschler]{neff2017generative}
Thomas Neff, Christian Payer, Darko {\v{S}}tern, and Martin Urschler.
\newblock Generative adversarial network based synthesis for supervised medical
  image segmentation.
\newblock In \emph{OAGM \& ARW Joint Workshop 2017 on “Vision, Automation \&
  Robotics”}. Verlag der Technischen Universit{\"a}t Graz, 2017.

\bibitem[Worrall et~al.(2016)Worrall, Wilson, and
  Brostow]{worrall2016automated}
Daniel~E Worrall, Clare~M Wilson, and Gabriel~J Brostow.
\newblock Automated retinopathy of prematurity case detection with
  convolutional neural networks.
\newblock In \emph{International Workshop on Large-Scale Annotation of
  Biomedical Data and Expert Label Synthesis}, pages 68--76. Springer, 2016.

\bibitem[Chen et~al.(2015)Chen, Xu, Wong, Wong, and Liu]{chen2015glaucoma}
Xiangyu Chen, Yanwu Xu, Damon Wing~Kee Wong, Tien~Yin Wong, and Jiang Liu.
\newblock Glaucoma detection based on deep convolutional neural network.
\newblock In \emph{Engineering in Medicine and Biology Society (EMBC), 2015
  37th Annual International Conference of the IEEE}, pages 715--718. IEEE,
  2015.

\bibitem[Frid-Adar et~al.(2018)Frid-Adar, Diamant, Klang, Amitai, Goldberger,
  and Greenspan]{frid2018gan}
Maayan Frid-Adar, Idit Diamant, Eyal Klang, Michal Amitai, Jacob Goldberger,
  and Hayit Greenspan.
\newblock Gan-based synthetic medical image augmentation for increased cnn
  performance in liver lesion classification.
\newblock \emph{arXiv preprint arXiv:1803.01229}, 2018.

\bibitem[Schawinski et~al.(2017)Schawinski, Zhang, Zhang, Fowler, and
  Santhanam]{schawinski2017generative}
Kevin Schawinski, Ce~Zhang, Hantian Zhang, Lucas Fowler, and Gokula~Krishnan
  Santhanam.
\newblock Generative adversarial networks recover features in astrophysical
  images of galaxies beyond the deconvolution limit.
\newblock \emph{Monthly Notices of the Royal Astronomical Society: Letters},
  467\penalty0 (1):\penalty0 L110--L114, 2017.

\bibitem[Lawrence(2004)]{lawrence2004accuracy}
Mary~Gilbert Lawrence.
\newblock The accuracy of digital-video retinal imaging to screen for diabetic
  retinopathy: an analysis of two digital-video retinal imaging systems using
  standard stereoscopic seven-field photography and dilated clinical
  examination as reference standards.
\newblock \emph{Transactions of the American Ophthalmological Society},
  102:\penalty0 321, 2004.

\bibitem[Massin et~al.(2003)Massin, Erginay, Ben~Mehidi, Vicaut, Quentel,
  Victor, Marre, Guillausseau, and Gaudric]{massin2003evaluation}
P~Massin, A~Erginay, A~Ben~Mehidi, E~Vicaut, G~Quentel, Z~Victor, M~Marre,
  PJ~Guillausseau, and A~Gaudric.
\newblock Evaluation of a new non-mydriatic digital camera for detection of
  diabetic retinopathy.
\newblock \emph{Diabetic medicine}, 20\penalty0 (8):\penalty0 635--641, 2003.

\bibitem[Szab{\'o} et~al.(2015)Szab{\'o}, Fiedler, Somogyi, Somfai,
  B{\'\i}r{\'o}, {\"O}lvedy, Hargitai, and N{\'e}meth]{szabo2015telemedical}
Dorottya Szab{\'o}, Orsolya Fiedler, Anik{\'o} Somogyi, G{\'a}bor~M{\'a}rk
  Somfai, Zsolt B{\'\i}r{\'o}, Veronika {\"O}lvedy, Zs{\'o}fia Hargitai, and
  J{\'a}nos N{\'e}meth.
\newblock Telemedical diabetic retinopathy screening in hungary: a pilot
  programme.
\newblock \emph{Journal of telemedicine and telecare}, 21\penalty0
  (3):\penalty0 167--173, 2015.

\bibitem[Abdellaoui et~al.(2016)Abdellaoui, Marrakchi, Benatiya, and
  Tahri]{abdellaoui2016screening}
M~Abdellaoui, M~Marrakchi, IA~Benatiya, and H~Tahri.
\newblock Screening for diabetic retinopathy by non-mydriatic retinal camera in
  the region of fez.
\newblock \emph{Journal francais d'ophtalmologie}, 39\penalty0 (1):\penalty0
  48--54, 2016.

\bibitem[Siu et~al.(1998)Siu, Ko, Wong, Chan, et~al.]{siu1998effectiveness}
SC~Siu, TC~Ko, KW~Wong, WN~Chan, et~al.
\newblock Effectiveness of non-mydriatic retinal photography and direct
  ophthalmoscopy in detecting diabetic retinopathy.
\newblock \emph{Hong Kong Medical Journal}, pages 367--370, 1998.

\bibitem[Chow et~al.(2006)Chow, Aiello, Cavallerano, Katalinic, Hock, Tolson,
  Kirby, Bursell, and Aiello]{chow2006comparison}
Sing-Pey Chow, Lloyd~M Aiello, Jerry~D Cavallerano, Paula Katalinic, Kristen
  Hock, Ann Tolson, Rita Kirby, Sven-Erik Bursell, and Lloyd~Paul Aiello.
\newblock Comparison of nonmydriatic digital retinal imaging versus dilated
  ophthalmic examination for nondiabetic eye disease in persons with diabetes.
\newblock \emph{Ophthalmology}, 113\penalty0 (5):\penalty0 833--840, 2006.

\bibitem[Wong and Bressler(2016)]{wong2016artificial}
Tien~Yin Wong and Neil~M Bressler.
\newblock Artificial intelligence with deep learning technology looks into
  diabetic retinopathy screening.
\newblock \emph{Jama}, 316\penalty0 (22):\penalty0 2366--2367, 2016.

\bibitem[Oke et~al.(2016)Oke, Stratton, Aldington, Stevens, and
  Scanlon]{oke2016use}
JL~Oke, IM~Stratton, SJ~Aldington, RJ~Stevens, and Peter~H Scanlon.
\newblock The use of statistical methodology to determine the accuracy of
  grading within a diabetic retinopathy screening programme.
\newblock \emph{Diabetic Medicine}, 33\penalty0 (7):\penalty0 896--903, 2016.

\end{thebibliography}

\end{document}